\documentclass[twoside,11pt]{article}
\usepackage{jair, rawfonts}

\usepackage[style=apa,natbib=true,doi=false,url=false]{biblatex}
\AtEveryBibitem{
\ifentrytype{inproceedings}{
 \clearlist{address}
 \clearlist{publisher}
 \clearname{editor}
 \clearlist{organization}
 \clearfield{url}  
 \clearfield{doi}  
 \clearfield{pages}  
 \clearlist{location}
 \clearlist{month}
 }{}
 
 \ifentrytype{article}{
 \clearlist{address}
 \clearlist{publisher}
 \clearname{editor}
 \clearlist{organization}
 \clearfield{url}  
 \clearfield{doi}  
 \clearlist{location}
 }{}
 }
\usepackage{booktabs}
\usepackage{bm}
\usepackage{multirow}
\usepackage{pdflscape}
\usepackage{amssymb}
\usepackage{amsmath}
\usepackage{quoting}
\usepackage{cleveref}
\usepackage{siunitx}
\usepackage{subcaption}
\usepackage{enumitem}

\usepackage{tikz}
\usetikzlibrary{arrows.meta, positioning, shapes.geometric}

\ShortHeadings{A Holistic Assessment of the Reliability of Machine Learning Systems}
{Corso, Karamadian, Valentin, Cooper \& Kochenderfer}
\firstpageno{1}

\begin{document}

\title{A Holistic Assessment of the Reliability of \\ Machine Learning Systems}

\author{\name Anthony Corso \email acorso@stanford.edu \\
       \addr Aeronautics and Astronautics,
       Stanford University,\\
       Stanford, CA 94305, USA
       \AND
       \name David Karamadian \email dk11@alumni.stanford.edu \\
       \addr Statistics,
       Stanford University,\\
       Stanford, CA 94305, USA
       \AND
       \name Romeo Valentin \email romeov@stanford.edu \\
       \addr Aeronautics and Astronautics,
       Stanford University,\\
       Stanford, CA 94305, USA
       \AND
       \name Mary Cooper \email marycoop@alumni.stanford.edu \\
       \addr Aeronautics and Astronautics,
       Stanford University,\\
       Stanford, CA 94305, USA
       \AND
       \name Mykel J.\ Kochenderfer \email mykel@stanford.edu \\ \addr Aeronautics and Astronautics,
       Stanford University,\\
       Stanford, CA 94305, USA}


\maketitle

\begin{abstract}
    As machine learning (ML) systems increasingly permeate high-stakes settings such as healthcare, transportation, military, and national security, concerns regarding their reliability have emerged. Despite notable progress, the performance of these systems can significantly diminish due to adversarial attacks or environmental changes, leading to overconfident predictions, failures to detect input faults, and an inability to generalize in unexpected scenarios. This paper proposes a holistic assessment methodology for the reliability of ML systems. Our framework evaluates five key properties: in-distribution accuracy, distribution-shift robustness, adversarial robustness, calibration, and out-of-distribution detection. A reliability score is also introduced and used to assess the overall system reliability. To provide insights into the performance of different algorithmic approaches, we identify and categorize state-of-the-art techniques, then evaluate a selection on real-world tasks using our proposed reliability metrics and reliability score. Our analysis of over \num{500} models reveals that designing for one metric does not necessarily constrain others but certain algorithmic techniques can improve reliability across multiple metrics simultaneously. This study contributes to a more comprehensive understanding of ML reliability and provides a roadmap for future research and development.
\end{abstract}

\section{Introduction}

Rapid progress in the capabilities of machine learning (ML) technology has prompted its nascent or proposed use in high-stakes settings such as transportation~\citep{ma2020artificial,Banavar2020applications}, healthcare~\citep{yu2018artificial,jiang2017artificial}, military~\citep{RR-3139-1-AF}, and national security~\citep{sayler2020artificial}. Despite this progress, ML components still suffer from various reliability issues, which compromise the ability for the system to perform its intended function during deployment. Adversarial or natural environmental changes can result in significant drops in performance, while poor calibration of uncertainty estimates can lead to overconfident predictions even in scenarios where accuracy is compromised. In the extreme case, ML components fail to detect input faults that should cause them to cease operation. So how should we evaluate ML models?

Recent guidance and regulation on artificial intelligence (AI) systems~\citep{oecd2019recommendation,nist2022airmf,european2021laying,order13960,dod2023autonomy}, which are typically enabled by ML technology, have outlined several key requirements for ensuring their safety, fairness, and security. While guidance and regulations vary, we summarize many of the key points addressed as follows. AI systems must have a well-defined objective and produce accurate, reliable outputs that generalize well and be robust to new scenarios. In the event of unexpected scenarios, AI systems must fail gracefully and minimize harm. To ensure safety, AI systems should undergo rigorous validation to identify failure modes and corresponding harms and be regularly audited for performance. To promote fairness, AI systems must be free from bias in data, algorithms, and human operators. Additionally, they should be secure, resilient, and protect privacy, through robustness against adversarial attacks and protection against sensitive data exfiltration. The design choices and decisions made by AI systems must be transparent and explainable, promoting trust among users. Lastly, there should be an accountable party to take responsibility when a bad outcome is realized. 

While some of these requirements need to be addressed at an organizational level, such as appropriate selection of objectives, auditing, transparency, and accountability, and others are addressed by traditional engineering safeguards such as data privacy and system-level security, the remaining requirements boil down to properties of the ML components themselves. Accuracy is a commonly used training objective and evaluation criterion for ML components, while generalization and robustness can be assessed through accuracy on diverse datasets that include distribution shifts, which are variations of the input data not captured by the training data. Adversarial robustness, measured by the ease with which performance is degraded by manipulating its inputs, is an important aspect of security. The ability of an AI system to fail gracefully depends on the uncertainty calibration and fault-detection capabilities of its ML components. These properties have been extensively studied, leading to advancements in improving each property in isolation, but often without measuring the effect on others. To build reliable AI systems, we need to evaluate the reliability of ML components holistically, considering all relevant metrics together. A holistic approach will provide a clearer understanding of progress toward building AI systems that we can safely deploy.

To that end, we propose a holistic assessment methodology of the reliability of machine learning systems. Guided by emerging regulations, we select five properties that are important for building reliable ML systems: 
\begin{enumerate}
    \item \textbf{In-distribution (ID) accuracy}: The accuracy on unseen data that comes from the same distribution as the training data. 
    \item \textbf{Distribution-shift (DS) robustness}: The ability to generalize to data that comes from a different distribution than the training data.
    \item \textbf{Adversarial robustness}: The ability to generalize to data that has been adversarially manipulated to degrade model performance. 
    \item \textbf{Calibration}: The accuracy of uncertainty estimates of the systems. 
    \item \textbf{Out-of-distribution (OOD) detection}: The ability to flag inputs that are out of the system's operational domain. 
\end{enumerate}
We identify quantitative metrics for each property and use them to construct a property score ranging from \num{0} to \num{1}. A final holistic reliability (HR) score is obtained by averaging the five property scores. We use this methodology to evaluate the many recent advances in building highly reliable ML systems. We survey the literature to identify state-of-the-art (SOTA) approaches to building highly accurate, robust, secure, calibrated, and fault-tolerant ML components. We group approaches based on where they fall within the ML system design cycle, which includes training data, representation learning, training objective, model architecture, model selection, post-hoc calibration, and run-time monitoring. We then select a representative set of approaches and use them to train models on three real-world tasks. These models are combined with other models from prior research trained on the same tasks, and each is evaluated according to the aforementioned reliability metrics and compared.

Our analysis of over \num{500} models reveals several key insights into the reliability of ML models. First, we find that these five reliability metrics are largely independent of each other, with only a few exceptions that appear to be dataset specific. This finding is consistent with the hypothesis that deep neural networks are underspecified~\citep{d2020underspecification}, meaning that designing to one metric does not constrain others due to the high number of degrees of freedom of modern models. Second, we find that some algorithmic approaches improve reliability across multiple metrics simultaneously. For example, we find that additional training data and model ensembling and fine-tuning from large pre-trained models all improve the holistic reliability score.

To summarize, in this work we make the following contributions:
\begin{itemize}
\item We survey the literature on building reliable ML systems and identify and categorize SOTA approaches. 
\item We propose a new methodology for the holistic assessment of the reliability of ML systems that combines metrics related to in-distribution performance, distribution-shift robustness, adversarial robustness, calibration, and out-of-distribution detection. 
\item We evaluate a representative set of SOTA ML techniques on a variety of real-world datasets. Through these experiments, we identify several technical approaches that improve the holistic reliability of ML systems independent of tasks. 
\end{itemize}

\section{A Survey of Reliable Machine Learning}
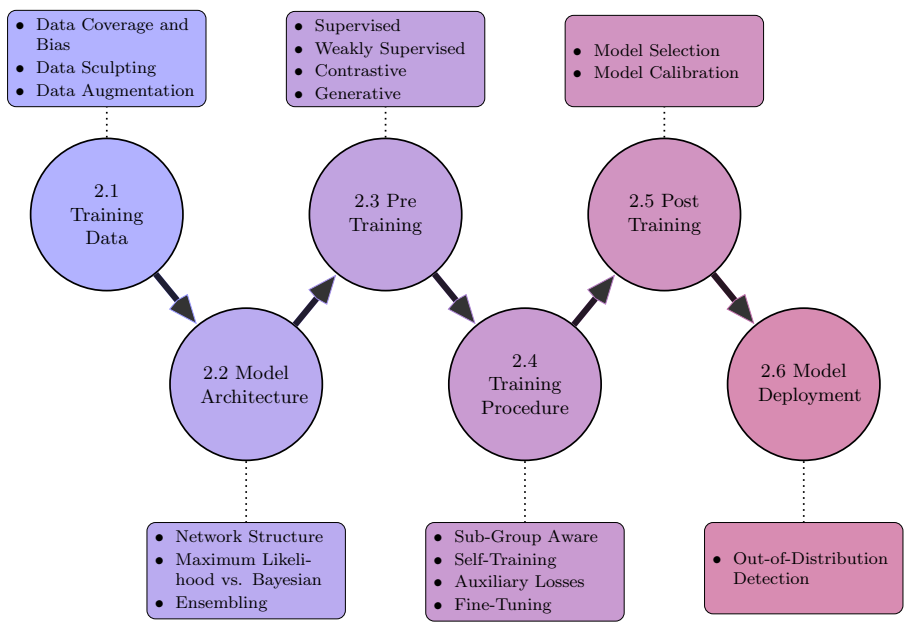
\begin{figure}
    \centering
    \resizebox{0.85\textwidth}{!}{ \definecolorseries{customseries}{rgb}{last}{blue!50!white}{purple!80!white}
\resetcolorseries[6]{customseries}

\colorlet{c1}{customseries!!+}
\colorlet{c2}{customseries!!+}
\colorlet{c3}{customseries!!+}
\colorlet{c4}{customseries!!+}
\colorlet{c5}{customseries!!+}
\colorlet{c6}{customseries!!+}

\begin{tikzpicture}[
  node distance=1cm and 0.5cm,
  myarrow/.style={
    line width=0.02mm, draw=#1, double=black, double distance=1mm, shorten >= 0.3mm,
    -{Stealth[inset=-1pt, length=13pt,  width=90pt, angle'=45, fill=black]},
    opacity=0.8,
  },
  mynode/.style={
    circle, fill=#1, fill opacity=0.6, text opacity=1.0,
    draw=black, thick, minimum size=2.2cm, minimum width = 2.5cm, align=center,
    text width=1.5cm, font=\footnotesize},
  mybox/.style={
    rectangle, minimum size = 1.5cm, align=center,
    fill=#1, fill opacity=0.6, text opacity=1.0,
    draw=black, text width=3cm, font=\scriptsize, rounded corners},
  mysmallbox/.style={
    rectangle, draw=black, very thin, align=center, fill=#1, fill opacity=0.6,
    text opacity=1.0,  text width=2cm, font=\scriptsize},
]

\node[mynode=c1] (step1) at (1,-1) {\ref{sec:training_data} Training Data};
\node[mynode=c2, below right=of step1] (step2) {\ref{sec:model_architecture} Model \\ Architecture};
\node[mynode=c3, above right=of step2] (step3) {\ref{sec:pretraining} Pre Training};
\node[mynode=c4, below right=of step3] (step4) {\ref{sec:training_procedure} Training Procedure};
\node[mynode=c5, above right=of step4] (step5) {\ref{sec:post_training} Post Training};
\node[mynode=c6, below right=of step5] (step6) {\ref{sec:model_deployment} Model \\ Deployment};

\draw[myarrow=c1] (step1) -- (step2);
\draw[myarrow=c2] (step2) -- (step3);
\draw[myarrow=c3] (step3) -- (step4);
\draw[myarrow=c4] (step4) -- (step5);
\draw[myarrow=c5] (step5) -- (step6);

\setlist[itemize]{leftmargin=*, topsep=0pt, itemsep=-1ex} 
\node[mybox=c1, above=0.5 of step1, text width=3cm] (box1) {
    \begin{minipage}{3cm}
    \begin{itemize}
    \item Data Coverage and Bias
    \item Data Sculpting
    \item Data Augmentation
    \end{itemize}
    \end{minipage}
};

\node[mybox=c2, below=of step2] (box2) {
    \begin{minipage}{3cm}
    \begin{itemize}
    \item Network Structure
    \item Maximum Likelihood vs. Bayesian
    \item Ensembling
    \end{itemize}
    \end{minipage}
};
\node[mybox=c3, above=0.5 of step3] (box3) {
    \begin{minipage}{3cm}
    \begin{itemize}
    \item Supervised
    \item Weakly Supervised
    \item Contrastive
    \item Generative
    \end{itemize}
    \end{minipage}
};
\node[mybox=c4, below=of step4] (box4) {
    \begin{minipage}{3cm}
    \begin{itemize}
    \item Sub-Group Aware
    \item Self-Training
    \item Auxiliary Losses
    \item Fine-Tuning
    \end{itemize}
    \end{minipage}
};
\node[mybox=c5, above=0.5 of step5] (box5) {
    \begin{minipage}{3cm}
    \begin{itemize}
    \item Model Selection
    \item Model Calibration
    \end{itemize}
    \end{minipage}
};
\node[mybox=c6, below=of step6] (box6) {
    \begin{minipage}{3cm}
    \begin{itemize}
    \item Out-of-Distribution Detection
    \end{itemize}
    \end{minipage}
};


\draw[dotted, thick] (step1) -- (box1);
\draw[dotted, thick] (step2) -- (box2);
\draw[dotted, thick] (step3) -- (box3);
\draw[dotted, thick] (step4) -- (box4);
\draw[dotted, thick] (step5) -- (box5);
\draw[dotted, thick] (step6) -- (box6);


\end{tikzpicture}}
    \caption{A categorization of approaches for reliable ML.}
    \label{fig:surveydiagram}
\end{figure}

We organize approaches for building more reliable ML systems into where they are applied during the design of a ML component. We summarize our categorization in \cref{fig:surveydiagram}, and expand in further detail in the following sections. We focus on systems that are trained offline and then deployed, rather than systems that adapt online, which we leave for future work. Systems that adapt online are more difficult to assess because they need to either 1) be reassessed each time they change, or 2) come with safety or reliability guarantees of the adaptation process itself. 

\subsection{Training Data}
\label{sec:training_data}

 The quality, diversity, and representativeness of training data can significantly impact the capabilities and generalization of ML systems. Adequate and diverse training data allows models to learn robust patterns, capture complex relationships, and handle various real-world scenarios. Insufficient or biased training data can lead to suboptimal performance, limited generalization, and potential ethical concerns. Therefore, ensuring the availability of high-quality, diverse, and unbiased training data is crucial for developing effective and fair ML systems~\citep{liang2022advances}.

\paragraph{Data Coverage and Bias} Selecting an initial subset of data that is representative of the entire dataset is a critical step in creating a reliable ML model. The choice of dataset, which should reflect the model's intended use case, plays a crucial role in the performance and robustness of the model. \citet{liang2022advances} advocate for an iterative data design process, typically starting with a pilot dataset to build a prototype model and then collecting more data to correct observed errors or expand the model's scope. \citet{cherti2022reproducible} underscores the importance of dataset size, finding that larger datasets generally enhance the performance of large pre-trained models. However, \citet{nguyen2022quality} demonstrates that the content of the pre-training dataset significantly affects the learning of robust models, reinforcing the necessity for quality and diversity in the data. Despite this, biases or insufficient data coverage can corrupt initial dataset selection, potentially leading to issues when the ML model is deployed, as seen in healthcare applications~\citep{obermeyer2019dissecting}, facial recognition models~\citep{buolamwini2018gender}, natural language processing~\citep{bolukbasi2016man,caliskan2017semantics}, and in many other applications~\citep{mehrabi2021survey}. Therefore, throughout this process, thorough documentation of data provenance is crucial, particularly noting biases and the socio-technical context of the data where applicable \citep{scheuerman2021datasets}.

\paragraph{Data Sculpting}  Data sculpting encompasses the selection, cleaning, annotation, and evaluation of data to enhance a model's reliability and generalizability \citep{liang2022advances}. Despite the importance of data annotation, it remains a source of significant error in AI models, with error rates such as 0.15\% for MNIST, 6\% for ImageNet, and 10.12\% for QuickDraw~\citep{northcutt2021pervasive}. Mitigation strategies include customized annotation approaches combining data programming~\citep{ratner2016data} and manual labeling or the adoption of weak supervision or unsupervised learning. The value of individual data points can be computed using Shapley values~\citep{ghorbani2019data}, which can be used to quantify data usefulness and filter out noisy, low-quality data.  Data sculpting also extends to evaluating model effectiveness and trustworthiness using techniques like data ablation. Data ablation is a process for detecting random correlations a model might use for predictions by testing how the model responds to missing key information, such as certain image details. For instance, researchers might obscure a polar bear in an image to see if the model predicts the animal's Arctic nativity based solely on the snowy background \citep{liang2022advances}.

\paragraph{Data Augmentation} Data augmentation is a widely used practice in deep learning that expands a fixed training set by generating additional samples through various techniques. It offers several benefits, including the expansion of available training data, promotion of model invariance, regularization through consistent predictions for augmented samples, and improved generalization~\citep{geiping2022much}. Early examples of data augmentation focused on simple geometric transformations~\citep{yaeger_aug}. Model-free methods~\citep{xu2023comprehensive} modify inputs through transformations~\citep{hendrycks2020augmix,cubuk2020randaugment}, combining datapoints~\citep{thulasidasan2019mixup,mintun2021interaction}, or using image-to-image networks~\citep{hendrycks2021many}. Model-based methods, such as incorporating synthetic examples of rare species~\citep{beery2020synthetic}, and policy-based methods such as adversarial training~\citep{madry2018towards} can also be used. These techniques have demonstrated performance improvements across various metrics, such as distribution-shift robustness~\citep{sagawa2022extending,hendrycks2018benchmarking,taori2020measuring}, expected calibration error (ECE)~\citep{thulasidasan2019mixup}, and out-of-distribution detection~\citep{thulasidasan2019mixup}. Adversarial training, employing techniques like Projected Gradient Descent (PGD), has also been successful in enhancing adversarial robustness against attacks, demonstrating significant improvements over baseline models~\citep{jeddiSimpleFinetuningAll2020,madry2018towards,ziegler2022adversarial}. While these techniques can enhance adversarial robustness, there is currently no known method that fully defends against adversarial attacks~\citep{xuAdversarialAttacksDefenses2020}.

\subsection{Model Architecture}
\label{sec:model_architecture}
Obtaining a high quality dataset is only one part of creating a good ML model. Another key design decision is the type of model to use. While many types of models exist, recently deep neural networks have been at the forefront. Deep neural networks are commonly used for processing text and image data but there are a large number of design considerations for choosing their architecture. In this section, we discuss the evolution of convolutional and transformer architectures, the advantages and challenges of Bayesian neural networks, and the power of ensembling diverse models.

\paragraph{Network Structure} Convolutional neural networks (CNNs) initially emerged as the dominant network structure for image recognition tasks, following their early success over dense layers. However, recent advancements have introduced transformer architectures like the Vision Transformer (ViT)~\citep{dosovitskiy2021an,bao2022beit}, Swin Transformer~\citep{liu2021swin}, and MaxViT~\citep{tu2022maxvit}, which have shown remarkable performance in language and image processing tasks, particularly with large datasets. Nevertheless, convolutional architectures such as ConvNeXt~\citep{liu2022convnet} and EfficientNet~\citep{tan2019efficientnet} have re-emerged to compete with transformers on the ImageNet benchmark.  When investigating model architectures, \citet{kornblith2019better} determined that there was a positive, but weak, correlation between ImageNet accuracy and accuracy on other unrelated tasks. However, \citet{fang2023progress} found that while earlier innovations on ImageNet show improvement on real-world datasets, the correlation does not exist for architectures developed after VGG~\citep{Simonyan2014very}. Instead, improvements have largely come from pre-training on large datasets, where it has been observed that model size is important for performance on downstream tasks~\citep{cherti2022reproducible}. 

\paragraph{Maximum Likelihood vs. Bayesian} Maximum likelihood estimation (MLE) is a common approach for training machine learning models, yielding point estimates for model parameters. In contrast, Bayesian machine learning provides a more natural representation of uncertainty by estimating a distribution over model parameters, which leads to a distribution over model outputs. Exact posterior updates for neural network parameters are intractable so approximations are made. Dropout~\citep{gal2016dropout}, a regularization technique that involves randomly zeroing out selected neuron weights, approximates Bayesian inference and can be applied at test time for uncertainty quantification. Markov chain Monte Carlo (MCMC) techniques have also been used but rarely achieve strong predictive performance due to the difficulty in estimating high-dimensional posterior distributions. \citet{hernandez2020improving} combined dropout with Hamiltonian Monte Carlo to obtain improved MCMC results. \citet{Mukhoti2018importance} conducted a comprehensive evaluation of approximate Bayesian inference techniques and identified dropout as a strong contender, though only relatively small datasets were used in the evaluation.

\paragraph{Ensembling} 
Ensembling involves aggregating the outputs or weights of multiple ML models into a single output or single model, which can then be used for the task at hand. Various ensembling techniques have been used to boost model performance across a number of metrics, including accuracy~\citep{wortsman2022model}, robustness~\citep{wortsman2022robust} and uncertainty quantification, where ensembles are used as an approximation to Bayesian Neural Networks~\citep{lakshminarayanan2017simple}. In the context of fine-tuning large pre-trained models, \citet{wortsman2022model} demonstrate that averaging the weights of multiple models fine-tuned with varying hyperparameter configurations can lead to improvements in both accuracy and robustness. Their approach with a ViT-G model led to a new state-of-the-art ImageNet accuracy of 90.94\%. Additionally, with large pre-trained models, \citet{wortsman2022robust} highlight the benefits of ensembling the weights of the zero-shot model and fine-tuned versions of the model. Doing so leads to large accuracy improvements under distribution shifts while preserving high accuracy on the distribution targeted by fine-tuning. 

When aggregating the outputs of multiple models into an ensemble, it has been widely shown that good ensembles are comprised of models that are both accurate and diverse, making errors independent of each other \citep{wen2020batchensemble, opitz1999popular, perrone1995networks,lee2016stochastic}, a feature that deep learning systems may often lack~\citep{mania2019model,mania2020classifier}. \citet{gontijo2021no} analyze the latest high-performing models, which are trained with diverse optimization objectives on different large-scale datasets. Their study and those of others~\citep{andreassen2021evolution} demonstrate that diverse training methodologies create models that specialize in different subdomains of the data and produce increasingly uncorrelated errors. These diverse models ensemble more efficiently, for example, ensembling a ResNet-50 model (ImageNet accuracy of 76.5\%) with its most different counterpart, ALIGN-ZS (ImageNet accuracy of 75.5\%) results in a 7\% boost in accuracy. 

\subsection{Pre-Training}
\label{sec:pretraining}

Pre-training is the process of training an ML component on datasets or tasks that are not necessarily related to the final downstream task. Often the goal in pre-training is to learn a representation of the data that eases the learning of the downstream task. A good representation~\citep{bengio2013representation} typically maps input data into a well-defined, low-dimensional manifold that represents the causal factors of the data-generating process. A good representation may improve each of the reliability metrics studied in this work. We overview four types of pre-training: supervised, weakly-supervised, contrastive, and generative. 

\paragraph{Supervised Pre-Training} Supervised pre-training involves training a model on an auxiliary dataset that is different from the primary task. It was initially used to enhance the performance of convolutional neural networks on tasks when there was limited training data~\citep{girshick2014rich,agrawal2014analyzing}. For example, pre-training on ImageNet led to a more than 40\% improvement in mean average precision (mAP) for the PASCAL VOC object detection task, with larger improvements for larger models~\citep{agrawal2014analyzing}. In a systematic study of pre-training from ImageNet,~\citet{kornblith2019better} found that ImageNet accuracy was a good predictor of downstream task performance both for logistic regression on the pre-trained featurizer as well as when fine-tuning the model. They found the largest improvements on datasets that were most similar to ImageNet, and lower improvements on datasets they deemed ``fine-grained'' such as those that make subtle distinctions between a large variety of cars or aircraft. Many of the convergence and performance benefits of pre-training were diminished when more data was available for the primary task. Supervised pre-training was also essential to the success of the Vision Transformer (ViT). The larger number of parameters and lack of built-in inductive biases of CNNs~\citep{li2021shapetexture,d2021convit}, mean that ViTs require more data.~\citet{dosovitskiy2021an} pre-trained their ViT architecture on the ImageNet-21k or JFT-300M datasets to get SOTA performance on ImageNet-1k, with the larger models seeing more improvement from the larger datasets.

\paragraph{Weakly Supervised Pre-Training} Weakly supervised pre-training involves converting unlabeled datasets into supervised learning tasks, allowing for the learning of effective representations. Examples of constructing supervised tasks from unsupervised data include predicting the relative location of two image patches~\citep{doersch2015unsupervised}, predicting image colors from grayscale~\citep{larsson2016learning}, and counting objects~\citep{noroozi2017representation}. Additionally, \citet{mahajan2018exploring,singh2022revisiting} demonstrated an effective form of weak supervision by constructing a dataset of 3.5 billion images from Instagram, where image hashtags serve as the image labels. Their approach, known as Supervised Weakly through hashtAGs (SWAG), showcased strong transferability to new datasets and achieved zero-shot performance comparable to models trained on image-text pairs~\citep{radford2021learning,jia2021scaling}.

\paragraph{Contrastive Pre-Training} Contrastive learning is a self-supervised learning approach that aims to learn useful representations by maximizing agreement between positive samples and minimizing agreement between negative samples. Instance contrastive learning~\citep{he2020momentum,wu2018unsupervised,misra2020self,bachman2019learning} uses different augmented views of the same instance as positive examples and views of different instances as negative samples~\citep{he2020momentum,misra2020self}. Swapping Assignments between multiple Views (SwAV)~\citep{caron2020unsupervised} assigns each image view to a cluster, alleviating the memory and compute requirements of pairwise feature comparison in a batch. By enforcing consistency between codes from different augmentations of the same image using the cluster assignments, the model learns discriminative representations without directly comparing the image features. Contrastive Language-Image Pre-training (CLIP)~\citep{radford2021learning} and A Large-scale ImaGe and
Noisy-text embedding (ALIGN)~\citep{jia2021scaling} employ image and text pairs to train models, with loss functions specifically designed to enhance matching of correct pairs. The resulting models show substantial improvements in effective robustness on distribution-shifted variants of ImageNet~\citep{radford2021learning,jia2021scaling}. \citet{kotar2021contrasting} finds that contrastive pre-training on datasets similar to the downstream tasks leads to improved performance in that task and combining datasets does not provide the benefits of each.

\paragraph{Generative Pre-Training} Generative pre-training encompasses unsupervised approaches aimed at reconstructing synthetically removed parts of the input data. Original methods employed autoencoder architectures~\citep{masci2011stacked} trained with a reconstruction loss or generative adversarial networks~\citep{goodfellow2020generative} to reconstruct full data examples. Additional generative pre-training tasks involve partial reconstruction, such as in-painting masked regions of an image or predicting the next word in a sentence~\citep{brown2020language}. Generative pre-training has proven highly effective in natural language processing, enhancing both accuracy and robustness~\citep{brown2020language}. In computer vision, masked image modeling (MIM) has emerged as an effective generative task~\citep{bao2022beit,xie2022simMIM,he2022masked}. For instance, Masked Autoencoders achieved SOTA performance on ImageNet-1K and ImageNet-C through MIM~\citep{he2022masked}. Furthermore, Contrastive Captioners (CoCa) introduced by \citet{yu2022coca} combine a contrastive loss between image and text encoders with a generative loss for captioning, leading to new state-of-the-art results on ImageNet when fine-tuned~\citep{yu2022coca}.

\subsection{Training Procedure}
\label{sec:training_procedure}

The traditional ML objective is empirical risk minimization (ERM)~\citep{vapnik1991principles}, which minimizes a loss over a finite set of samples, typically with some form of stochastic gradient descent. Additional losses and optimization techniques have been proposed to improve one or more reliability metrics and are discussed below. 

\paragraph{Subgroup-Aware Training} Rather than only minimizing the empirical risk across the dataset, subgroup-aware training objectives incorporate domain knowledge. To use these approaches, the data must be categorizable into discrete subgroups. In classification problems, the label itself can be the group identifier. In other settings, the data may be augmented with additional metadata to identify subgroups. Distributionally robust optimization (DRO)~\citep{Sagawa2020Distributionally,duchi2021statistics,hu2018does} minimizes the worst-case performance in a subgroup, but requires stronger-than-normal regularization to ensure good generalization~\citep{Sagawa2020Distributionally}. Other approaches such as invariant risk minimization (IRM)~\citep{arjovsky2019invariant} and risk extrapolation~\citep{krueger2021out} promote risk-invariance across domains in order to produce more generalizable models. Correlation alignment (CORAL)~\citep{sun2016return,sun2016deep} methods try to align the second-order statistics of input features (extracted from deep neural networks in deep CORAL~\citep{sun2016deep}) between domains on labeled or unlabeled data. Domain-Adversarial Training of Neural Networks (DANN)~\citep{ganin2016domain}, promotes the learning of features that are discriminative for the primary task but are indiscriminative for the domain. Adaptive Feature Norm (AFN)~\citep{xu2019larger}, tries to ensure the magnitude of feature norms is the same across domains and demonstrates that when this is the case, models have better generalization. 

\paragraph{Self-Training on Unlabeled Data}
Self-training involves iteratively training a model on the labeled data initially available and then using this model to make predictions on unlabeled data, which is subsequently incorporated as ``pseudo-labeled" data for further training iterations~\citep{yarowsky-1995-unsupervised}. This iterative process of self-training with pseudo-labels helps harness the information present in unlabeled data, improving model performance and generalization. The PseudoLabels approach~\citep{lee2013pseudo} involves labeling unlabeled training data using the predictions of the current model. Several works have extended this idea by incorporating two models, a student and a teacher~\citep{xie2020self,pham2021meta,touvron2021training}. The teacher labels unlabeled data for the student, which is trained on input data that is typically noised through techniques like data augmentation or dropout. The Noisy Student method~\citep{xie2020self}, utilizing 300M unlabeled examples, achieved state-of-the-art results on in-distribution and robustness ImageNet benchmarks. FixMatch~\citep{sohn2020fixmatch} selectively uses the pseudo-labels that are predicted with high confidence and introduces a high level of noise during student training, showcasing remarkable effectiveness in low-labeled-data scenarios.

\paragraph{Auxiliary and Alternative Losses}
Auxiliary loss functions play a critical role in improving machine learning models by incorporating additional objectives into the training process. In the context of uncertainty quantification, \citet{kumar2018trainable} propose a loss function that directly optimizes model calibration during training and \citet{seo2019learning} include entropy regularization, both of which enhance the calibration of uncertainty estimates without the need for additional calibration steps.  For enhancing the smoothness of the loss landscape, \citet{foret2021sharpnessaware} introduce sharpness-aware minimization, a technique that aims to find minima in the loss landscape with uniformly low values. Models trained with this loss exhibit improved generalization performance across various tasks, providing a valuable tool for reducing overfitting and achieving better model behavior in practice. \citet{koh2020concept} propose a method that first predicts a set of human-interpretable concepts as intermediate representations before performing the final prediction task. By leveraging such concept-based representations, models can capture meaningful factors of variation in the data, enabling better interpretability and performance in complex tasks.


\citet{kilbertus2018generalization} argue that capturing the correct causal model is crucial for strong generalization properties. They motivate approaches that explicitly disentangle concepts or employ generative models, enabling better control and understanding of the underlying data distribution. For example, \citet{wang2017safer} introduce a generative classification approach based on a combination of variational autoencoders (VAEs) and generative adversarial networks (GANs), which demonstrates superior performance in tasks such as out-of-distribution detection. These approaches, however, have yet to scale to real-world datasets and tasks. 

\paragraph{Fine-Tuning} Recent advancements in fine-tuning have led to improvements in performance on not only in-distribution accuracy, but also domain-shift robustness and adversarial robustness. \citet{andreassen2021evolution} note how in-distibution accuracy versus domain-shifted accuracy typically follows a single linear trend, and models with a higher domain-shifted accuracy compared to this baseline exhibit ``effective robustness''. Through their investigations, they demonstrated that zero-shot pre-trained models, as well as those models that have been minimally fine-tuned, exhibit effective robustness, but this effective robustness is lost as the models are fine-tuned to convergence. \citet{wortsman2022robust} address this reduction in effective robustness from fine-tuning by ensembling the weights of the zero-shot and fine-tuned models; compared to standard fine-tuning, which achieved accuracies of 86.2\% on ImageNet ID and an average of 68.6\% across five distribution shifts, the ensembled model achieved accuracies of 87.1\% and 77.4\% on the same datasets. Fine-tuning is particularly sensitive to the optimization algorithm used, and \citet{kumar2022fine} develop new optimization strategies to improve the accuracy and robustness of fine-tuned models. Adversarial fine-tuning has been shown to achieve competitive adversarial robustness compared to adversarial training from scratch, saving substantial computational effort~\citep{jeddiSimpleFinetuningAll2020}. However, special care has to be taken to not overfit to the adversarial examples, and reduce the original performance, so careful learning rate scheduling was applied. 


\subsection{Post-Training}
\label{sec:post_training}

After training is complete, additional analysis and calibration may be required before a model can be deployed. We discuss algorithms for selecting trained models from a set, as well as calibrating models for more accurate uncertainty estimates. 

\paragraph{Model Selection} Selecting the most suitable model for deployment can be a challenging task, even when multiple models demonstrate similar performance on in-distribution test sets. Variations in other crucial properties, such as robustness, can exist among models trained using nearly identical procedures~\citep{gulrajani2021in,d2020underspecification}. In their exploration of domain generalization approaches, \citet{gulrajani2021in} find that using a validation set that includes all of the training domains outperforms leave-one-domain-out cross-validation. Additionally, \citet{d2020underspecification} discover that underspecification issues result in models exhibiting diverse robustness properties, which are not reflected solely in the test set performance. These findings underscore the importance of carefully considering various factors beyond conventional evaluation metrics when selecting models for deployment.

\paragraph{Uncertainty Calibration} Once a model has been trained, its uncertainty estimates can be modified without affecting the model accuracy. Overconfidence has been observed across deep learning architectures, leading to the development of techniques that can correct for it and even provide formal guarantees of the predictive uncertainty. \citet{guo2017calibration} introduce temperature scaling which is the best of their compared approaches to uncertainty quantification. For the $K$-class classification setting with logits $z_i$ and temperature $T$, the probability of the $i$th class is:
\begin{equation}
\mathbb{P}(y_i) = \frac{\exp(z_i/T)}{\sum_{k=1}^K \exp(z_k/T)}
\label{eq:temperatureScaling}
\end{equation}
The temperature is chosen to minimize the negative log-likelihood of a validation set. 
Another calibration procedure is conformal prediction~\citep{angelopoulos2023conformal}, which allows for the construction of prediction regions with provable guarantees from heuristic notions of uncertainty. However, without initially well-calibrated uncertainty estimates, the prediction regions are often too large to be useful~\citep{angelopoulos2021uncertainty}.

\subsection{Model Deployment}
\label{sec:model_deployment}

Once a ML system has been sufficiently tested and evaluated it may be deployed in a real-world setting. Deployment, however, is not a on-time event, but rather a continuous process of monitoring, updating, and improving models based on new data, feedback, and objectives. Regular assessment of a system's performance is essential since performance tends to degrade over time due to drifts in the data distribution~\citep{yao2022wildtime}.If performance drops below an acceptable level, the model should be updated with more recent data.

More dramatic shifts in the input distribution due to faults or other anomalies are detected in a process known as run-time monitoring. If the space of possible faults is known at design time, a system can be designed to look out for these faults and respond accordingly. To identify unforseen events, however, we must rely on approaches for out-of-distribution detection.

\paragraph{Out-of-Distribution Detection} Out-of-distribution (OOD) detection (also called anomaly detection or fault detection) refers to the task of identifying inputs to a ML model that come from a different distribution than the training data. Algorithms for OOD detection typically involve computing a score and using it with a predefined threshold for flagging OOD inputs. The algorithms primarily differ in the way the score is computed. A baseline approach to OOD detection is to use the max softmax value of the output as a score indicating the confidence of in-distribution, with the assumption that in-distribution inputs should be classified with higher confidence~\citep{hendrycks2017a}. \citet{pmlr-v162-hendrycks22a} found that using the max logit outperformed the max softmax, since information is lost due to the normalization of the softmax operation. \citet{NEURIPS2020_f5496252} uses an energy score that combines information from all class labels to better identify OOD inputs. \citet{liang2018enhancing} introduce ODIN (Out-of-Distribution detector for Neural networks) which computes the max softmax of an adversarially perturbed input. They observe that OOD samples have a higher sensitivity to such perturbations and therefore produce higher perturbed softmax scores. Other approaches rely on information from training such as typical gradients~\citep{sharma2021sketching}.

\section{Experimental Setup}
Given the wide range of approaches to building reliable ML models, we want to investigate the following two questions:
\begin{enumerate}
    \item How do models trained using different ML approaches perform on a variety of reliability metrics?
    \item How do different reliability metrics relate to each other?
\end{enumerate}

To answer these questions, we perform a large scale evaluation of machine learning models across a variety of reliability metrics. We use a combination of models trained for prior work and models trained for the present work to evaluate the relationship between machine learning methodology and reliability metrics. We also evaluate the relationship between reliability metrics by measuring the correlation between them. In the following sections, we describe the datasets, metrics, and models used in our assessment.

\subsection{Selection of Datasets}
A small number of benchmarks such as ImageNet, have become the de-facto standard in machine learning research, but it is not clear whether progress on those datasets translates to progress on other tasks~\citep{fang2023progress,kotar2021contrasting}. For that reason, we chose to use other publicly available datasets for evaluation. To choose the datasets, we used the following criteria:
\begin{enumerate}
    \item The datasets should reflect a real-world task of practical importance.
    \item The datasets should include data from real-world distribution shifts.
    \item The datasets should be image-based to make the use of pre-trained models feasible.
    \item The task should be binary or multi-class classification for the ease of using open source adversarial attacks and defining a calibration metric.
\end{enumerate}
While there have been a number of benchmarks dedicated to measuring distribution-shift generalization~\citep{koh2021wilds,sagawa2022extending,malinin2021shifts,malinin2022shifts,beery2019efficient,gulrajani2021in}, we found that the image classification tasks provided by the WILDS benchmark~\citep{koh2021wilds,sagawa2022extending} satisfied these requirements best due to the real-world nature of the tasks and the existence of models trained using a variety of approaches from prior work. We selected the following datasets for our analysis:
\begin{enumerate}
    \item \textbf{iWildCam}: A dataset for species recognition (\num{182} species labels) from camera-trap images. The data is annotated with the camera identification number that the image was taken from.  Each of the train, validation, and test datasets have a disjoint subset of identification numbers.
    \item \textbf{FMoW}: A dataset for land-use classification (\num{62} land-use labels) from satellite images. The data is annotated with the year the image was taken and the geographical region of the image. The training dataset is composed of images from before 2013, the validation set is from 2015--2016, and the test set consists of images from 2016 onward. 
    \item \textbf{Camelyon17}: A dataset for the binary classification of metastatic breast cancer from patches of an image recorded from lymph node tissue under a microscope. The data is annotated with the hospital that the slide was prepared and imaged in. The training dataset contains data from three hospitals while the validation and test sets each have data from a different hospital.
\end{enumerate}
See the WILDS benchmark for additional information on the datasets~\citep{koh2021wilds}.

\subsection{Selection of Evaluation Metrics}
Based on the literature review and emerging regulations, we have identified five important metrics related to the reliability of machine learning systems: in-distribution performance, distribution-shift robustness, adversarial robustness, calibration, and out-of-distribution detection. Each of these metrics can be measured in a variety of ways and in this subsection, we describe and justify our methods of evaluation. For each metric, we define a corresponding score $s \in [0,1]$ so that larger values are always preferable.

\paragraph{In-Distribution Performance} A primary assumption in statistical machine learning is that data are independently and identically distributed (iid)~\citep{hastie2009elements}. Datasets are typically split into a train set, used to fit a statistical model, a validation set, used to select hyperparameters, and a test set, used to evaluate the performance of the selected model. In classification settings, the primary metric of interest is accuracy, which is the fraction of correctly predicted labels. Precision, recall, and F1 score, which are computed on binary classification tasks and can be aggregated in the multi-class setting, are also important performance metrics. For regression tasks, the mean absolute error (MAE) or the mean squared error (MSE) are typical metrics of performance. To remain general, we indicate the performance of a ML model $m$ with respect to a dataset $\mathcal{D}$ as 
\begin{equation}
p = \text{performance}(m,\mathcal{D})\text{.}
\end{equation}
The in-distribution performance $p_{\rm ID}$ is measured on a held-out test set from the same distribution as the training data. We denote the performance score
\begin{equation}
s_{\rm ID} = p_{\rm ID}
\end{equation}
where in all of our experiments, $p_{\rm ID}$ is the top-1 classification accuracy. 

\paragraph{Distribution-Shift Robustness}

\begin{figure}
    \centering
    \includegraphics[width=\textwidth]{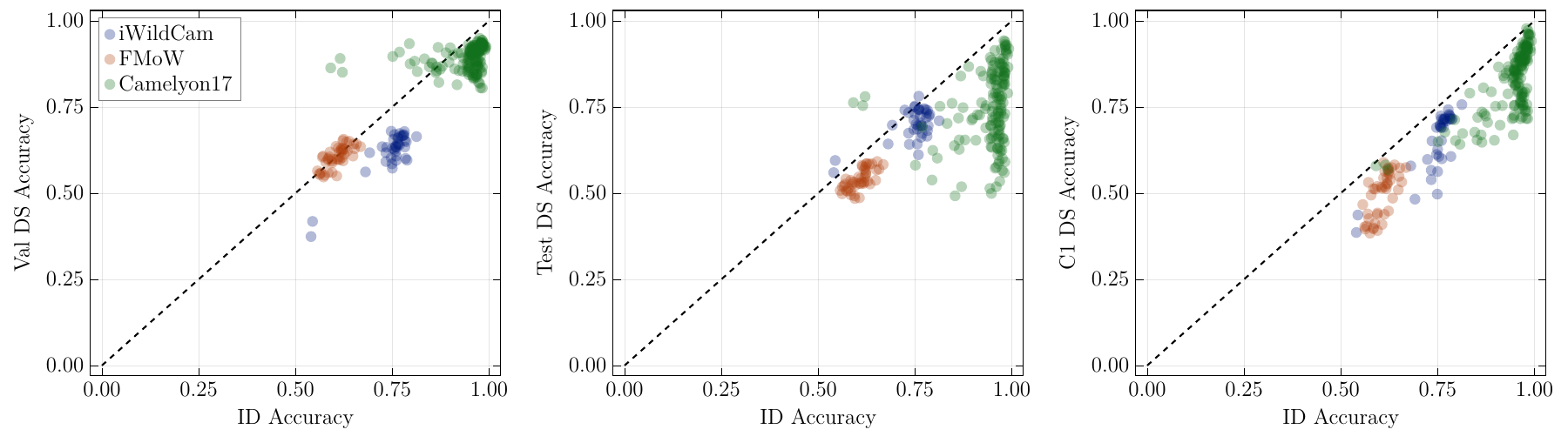}
    \caption{Performance of WILDS pre-trained models under various distribution shifts vs. in-distribution performance. The validation distribution shift and the test distribution shift come from the WILDS benchmark while C1 measures the performance drop from synthetic corruptions~\citep{hendrycks2018benchmarking} of strength 1.}
    \label{fig:performance_drop}
\end{figure}

When the iid assumption is violated, the performance of modern deep learning systems can drop significantly---far more than would be normal for a human~\citep{Geirhos2018generalization}. This effect has been demonstrated with synthetic distribution shifts such as common image corruptions and perturbations~\citep{hendrycks2018benchmarking,Geirhos2018generalization,mintun2021interaction}, as well as natural distribution shifts~\citep{taori2020measuring} arising from re-collecting data~\citep{recht2019imagenet}, changing data collection criteria~\citep{hendrycks2021many}, perturbations over time~\citep{gu2019using,shankar2021image}, or other real-world changes in the deployment environment~\citep{koh2021wilds,malinin2021shifts,malinin2022shifts,beery2019efficient}.  \Cref{fig:performance_drop} shows the performance of the WILDS pre-trained models subject to real-world shifts from the WILDS dataset (validation and test datasets, which each represent a different shift) as well as synthetic shifts from the standard set of image corruptions with corruption level 1 (C1)~\citep{hendrycks2018benchmarking}. 

While there exist no guarantees of performance under arbitrary distribution shifts, it has been shown empirically that there is sometimes strong correlation between in-distribution generalization and distribution-shift generalization~\citep{recht2019imagenet,gu2019using,miller2021accuracy,andreassen2021evolution}, a phenomenon referred to as accuracy on the line~\citep{miller2021accuracy}. We observe these linear trends in \cref{fig:performance_drop} for iWildCam and FMoW, but Camelyon17 is more highly varied, which has also been observed in prior work~\citep{miller2021accuracy}. The existence of these linear trends has lead researchers to measure not just the performance under a distribution shift but the effective robustness~\citep{andreassen2021evolution,wortsman2022robust,taori2020measuring} of a model, which is the amount the performance deviates from the value predicted by a linear model over the in-distribution generalization. In a similar vein, we separate robustness from performance of the model by measuring the ratio of performance under the distribution shift to the in-distribution performance, so a robustness value of \num{1} means the performance does not drop compared to the in-distribution performance.  

\begin{figure}
    \centering
    \includegraphics[width=0.6667\textwidth]{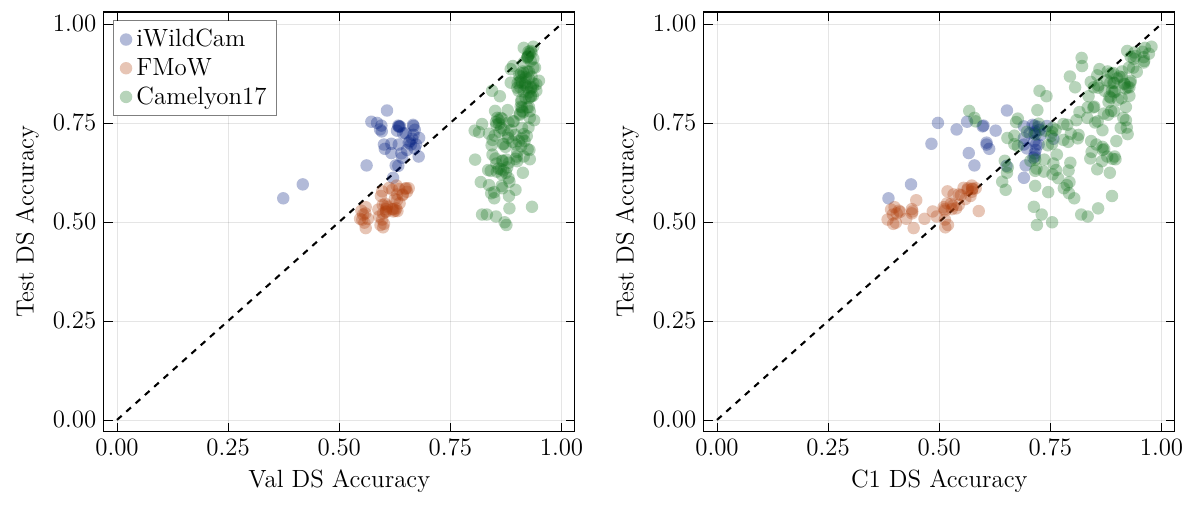}
    \caption{Comparing distribution-shift accuracy between various distribution shifts.}
    \label{fig:ds_performance_comparisons}
\end{figure}

For some datasets and machine learning approaches, however, the accuracy on the line trend is weaker, so it is still necessary to measure the distribution-shift performance on various distribution shifts.  The type and magnitude of distribution shifts one should evaluate a model on is context-dependent and should ideally reflect the deployment environment of the model. However, this is not always possible to know a priori so the question arises:
\emph{Does robustness to one distribution shift imply robustness to others?} To investigate this question, we compare the performance of the WILDS models to three different distribution shifts. Following the conventions of the WILDS datasets, ``val'' and ``test'' are different real-world shifts, while ``C1'' refers to a synthetic corruption shift~\cite{hendrycks2018benchmarking} of strength 1. The results shown in \cref{fig:ds_performance_comparisons} indicate that although there may be some correlation, it is weaker than the correlation to in-distribution generalization, an observation consistent with prior work~\citep{Geirhos2018generalization,mintun2021interaction}.  We, therefore, choose to evaluate model robustness on multiple distribution shifts and average the results. For $N$ distribution-shifted datasets $\{ \mathcal{D}_{\rm DS, 1}, \ldots, \mathcal{D}_{\rm DS, N} \}$, with model performance $p_i$ on the $i$th dataset, the distribution shift robustness score is
\begin{equation}
    s_{\rm DS} = \frac{1}{N} \sum_i \frac{p_i}{p_{\rm ID}}.
\end{equation}
In our experiments we average across the ``val'', ``test'', and ``C1'' distribution shifts. We choose averaging (instead of worse case) because we want the models with the highest score to do well on all three distribution shifts.

\paragraph{Adversarial Robustness}
Despite models being able to generalize to unseen samples from previously seen distributions, well-chosen yet small input perturbations can catastrophically change the model outputs~\citep{goodfellowExplainingHarnessingAdversarial2015,madry2018towards}.
Such ``adversarial attacks'' can be constructed in a wide variety of ways, for instance, by using a model's gradient information, or by transferring attacks from similar models. In the presence of an adversary, Adversarial Robustness then denotes the capability of a model to withstand adversarial attacks (with limited strength) for a fixed and finite set of data samples.
For example, one may measure adversarial robustness by computing the misclassification rate of a set of samples when subjected to adversarial attacks with a fixed maximum input perturbation.

In order to evaluate adversarial robustness, we follow the procedure of \citet{croceReliableEvaluationAdversarial2020a} and employ an ensemble of parameter-free adversarial attacks, consisting of two step-size free modifications of Projected Gradient Descent (APGD-CE and APGD-DLR), as well as Fast Adaptive Boundary Attack~\citep{croceMinimallyDistortedAdversarial2020} and Square Attack~\citep{andriushchenkoSquareAttackQueryEfficient2020}.
We note that this is the same approach as used by the popular ``RobustBench'' benchmark~\citep{croce2021robustbench}.
For the attack strength, we choose a maximum perturbation of \(\epsilon = 3/255\) in the \(L_\infty\)-norm, which is consistent with common choices for natural images.
In order to avoid conflating effects, we do not consider inputs from distributional shifts, and instead only evaluate on unseen in-distribution samples.

Then, if $p_{\rm ADV}$ is the performance in the presence of adversarial examples, we report the adversarial robustness score as the ratio in performance compared to the in-distribution performance
\begin{equation}
\label{eq:adv_rob}
    s_{\rm ADV} = \frac{p_{\rm ADV}}{p_{\rm ID}}.
\end{equation}

\paragraph{Calibration}
While it may not be possible for ML systems to be highly accurate in every situation, it is essential for these systems to accurately quantify their level of uncertainty (for a review see~\citet{abdar2021review}). Here we assess uncertainty quantification using the notion of calibration, where a model is calibrated if its confidence measure aligns with the error distribution of the outputs. A model can be both highly accurate and poorly calibrated, or conversely, highly inaccurate and well-calibrated. Several empirical measures exist for evaluating model calibration. For example, \citet{guo2017calibration} explore metrics such as the expected calibration error ($\mathit{ECE}$), which is computed by binning the model outputs and averaging the absoluted difference in the accuracy of each bin to the accuracy predicted by an identity function. \citet{guo2017calibration} also introduce  the maximum calibration error, which calculates the largest deviation across bins. They also discuss the negative log-likelihood, a metric also used by \citet{lakshminarayanan2017simple} and \citet{gal2016dropout}, as it is applicable to regression tasks in addition to classification tasks, though $\mathit{ECE}$ has also been extended to the regression setting~\citep{levi2022evaluating}. In this work, since we focus on classification tasks we use $\mathit{ECE}$. To convert $\mathit{ECE}$ to a calibration score, we normalize it by a maximum error $\mathit{ECE}_{\rm max}$ and subtract the result from \num{1}. We use $\mathit{ECE}_{\rm max}=0.5$ as this corresponds with the pathological case of a binary classifier that is correct 50\% of the time but always predicts with 100\% confidence.

\begin{figure}
    \centering
    \includegraphics[width=\textwidth]{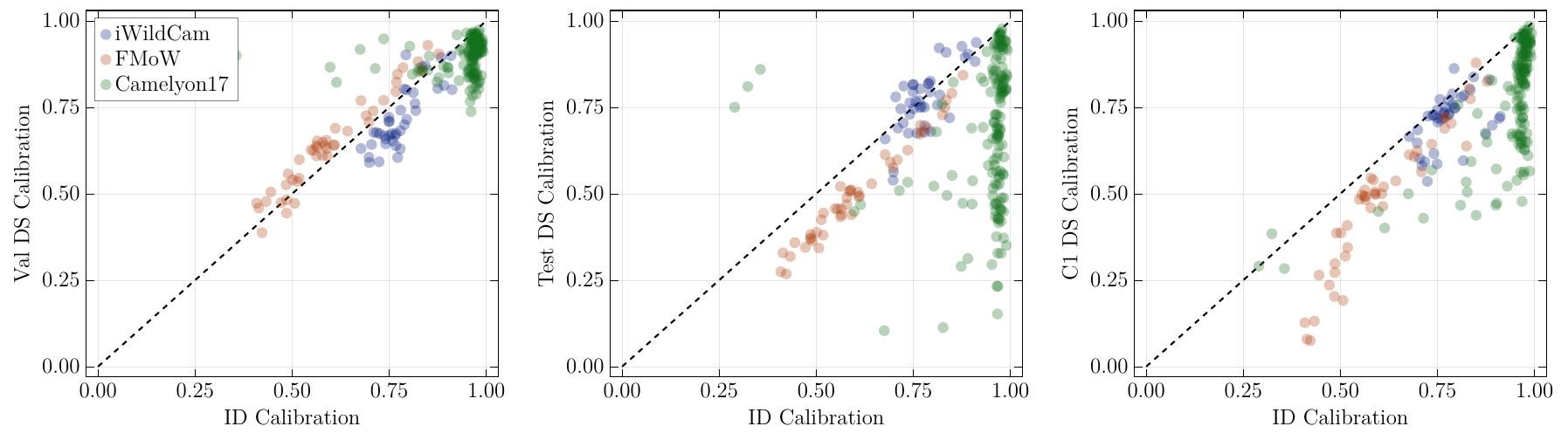}
    \caption{Calibration of WILDS pre-trained models under various distribution shifts vs. in-
distribution performance.}
    \label{fig:calibration_drop}
\end{figure}

\begin{figure}
    \centering
    \includegraphics[width=0.66667\textwidth]{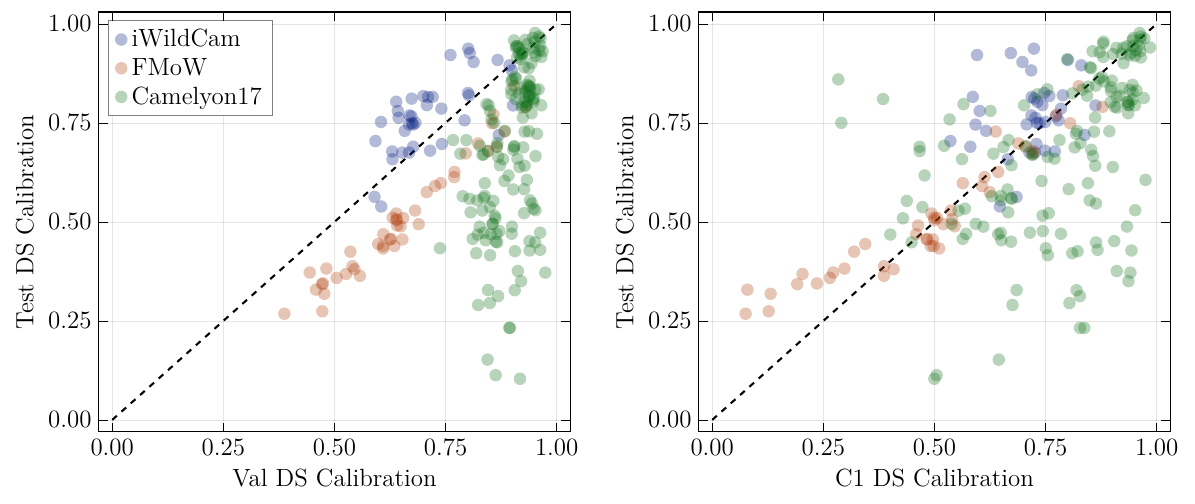}
    \caption{Comparing distribution-shift calibration between various distribution shifts.}
    \label{fig:ds_calibration_comparisons}
\end{figure}

Ideally, a model should be well-calibrated on both in-distribution data as well as on distribution-shifted data, even if the model performance decreases. In \cref{fig:calibration_drop}, we compare the calibration of the WILDS pre-trained models across a variety of datasets and find that the calibration of a model often drops under distribution shifts, and in some cases (iWildCam and FMoW) exhibits a significant correlation with the in-distribution calibration. For Camelyon17, however no strong correlation seems to exist. We also check whether calibration on one type of distribution shift implies calibration on others. \Cref{fig:ds_calibration_comparisons} shows that similarly to performance, the correlation is weaker between distribution shifts than it is to the in-distribution calibration. We therefore choose to compute the calibration score by averaging the calibration of the model across the in-distribution and distribution-shifted datasets, giving
\begin{equation}
    s_{\rm CAL} = 1 - \frac{1}{(N+1) \mathit{ECE}_{\rm max}} \left[ \mathit{ECE}_{\rm ID} + \sum_{i=1}^N \mathit{ECE}_i \right] \text{,}
\end{equation}
where $\mathit{ECE}_{\rm ID}$ is the expected calibration error on the in-distribution data and $\mathit{ECE}_i$ is measured on the $i$th distribution-shifted dataset. As for DS robustness, we average across the ``val'', ``test'', and ``C1'' distribution shift and use averaging instead of picking the worst case for the same reasons. 

\paragraph{Out-of-Distribution Detection}
The identification of out-of-distribution (OOD) inputs is a binary classification task and can therefore be measured according to a variety of metrics including precision, recall, and F1 score. These metrics, however, typically require the a priori definition of a threshold parameter, which would be set using validation data and a context-specific notion of the desirable precision-recall trade-off. To avoid selecting a threshold, we follow prior work in using the area under the receiver operator curve ($\mathit{AUROC}$) to assess the ability of a model to identify OOD inputs. 

To measure OOD detection we require datasets that are OOD. Ideally, these datasets should reflect the types of input faults that may occur for the machine learning model, but this may not be possible to know at the time of model evaluation. We therefore select $M$ diverse OOD datasets and combine them into a single OOD dataset
\begin{equation}
     \mathcal{D}_{\rm OOD} = \mathcal{D}_{\rm OOD}^{(1)} \cup  \mathcal{D}_{\rm OOD}^{(2)} \ldots \cup \mathcal{D}_{\rm OOD}^{(M)}.
\end{equation}
 For each task we evaluate from WILDS, we use the other image-classification tasks as well as a dataset of Gaussian noise and the RxRx1 dataset~\citep{koh2021wilds} as our OOD data. For example, when performing OOD detection on models trained on iWildCam, we compose the ID validation sets from Camelyon17, FMoW, RxRx1, and Gaussian noise to make up a single OOD dataset. 

\begin{figure}
  \centering
  \begin{subfigure}[t]{0.35\textwidth}
    \centering
    \includegraphics[width=\textwidth]{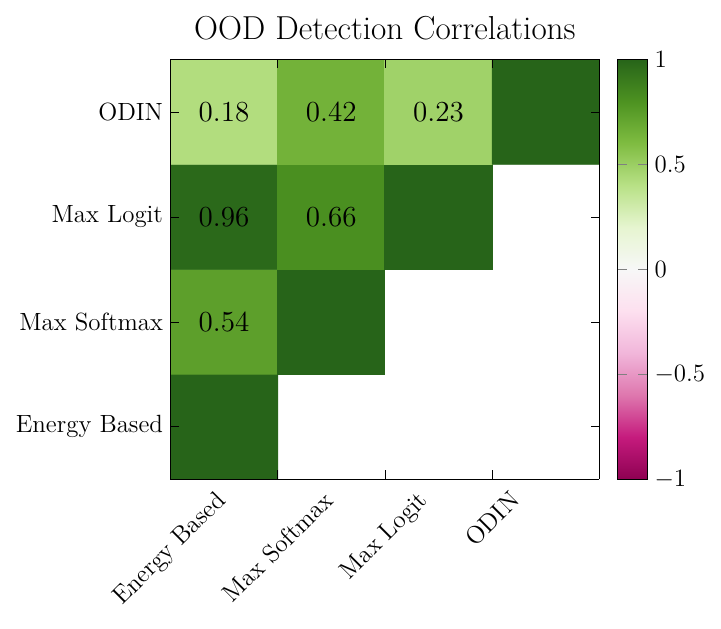}
    \caption{Correlation of $\mathit{AUROC}$ values for each pair of OOD detection algorithms. Color is Pearson correlation and number is $R^2$.}
    \label{fig:ood_detection_correlation}
  \end{subfigure}\hfill%
  \begin{subfigure}[t]{0.58\textwidth}
    \centering
    \includegraphics[width=\textwidth]{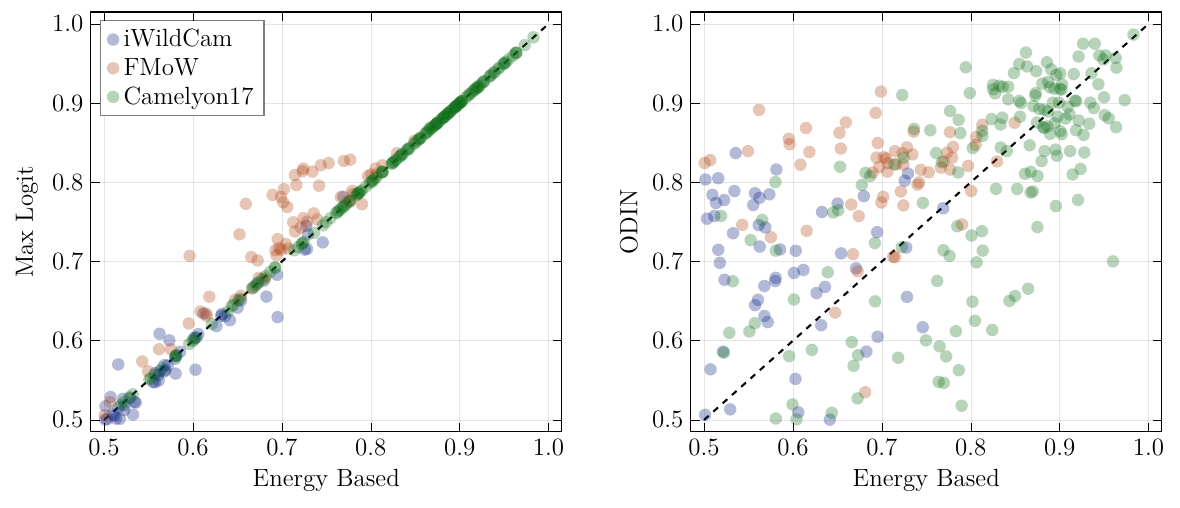}
    \caption{Comparison of the highest and lowest correlations between OOD detection algorithms.}
\label{fig:ood_detection_scatter}
  \end{subfigure}
  \caption{Comparing various OOD detection algorithms on models trained from each of the three datasets to check for consistency in detection strategies.}
  \label{fig:ood_evaluation}
\end{figure}

There are many approaches to outputting OOD detection scores from model outputs. We use the open source implementation\footnote{\url{https://github.com/kkirchheim/pytorch-ood}} of four different OOD detection algorithms (Max Logit~\citep{pmlr-v162-hendrycks22a}, Max Softmax~\citep{hendrycks2017a}, Energy-based~\citep{NEURIPS2020_f5496252}, and ODIN~\citep{liang2018enhancing}) and compare their $\mathit{AUROC}$. In \cref{fig:ood_detection_correlation}, we plot the Pearson correlation of each approach's $\mathit{AUROC}$ (combining all three datasets) and find that although there is some correlation between approaches, there can be substantial variability (we show scatter plot comparisons of the highest and lowest correlations in \cref{fig:ood_detection_scatter}). For this reason, we compute the final OOD detection score as the average of the $\mathit{AUROC}$ for each of $N_{\rm OOD}$ OOD detection algorithms
\begin{equation}
     s_{\rm OOD}=\frac{1}{N_{\rm OOD}} \sum_{i=1}^{N_{\rm OOD}} \mathit{AUROC}_i.
\end{equation}
We choose to average this metric under the hypothesis that a model that does well with multiple OOD detection algorithms has a representation that is well-suited to OOD detection in general, and should therefore receive a higher score.

\paragraph{Holistic Reliability Score} For many applications it is desirable for a model to score highly on all of the reliability metrics simultaneously. To measure overall reliability, we introduce the \emph{holistic reliability} score as the weighted average across properties. If we define $\mathbf{w} = [w_{\rm ID}, w_{\rm DS}, w_{\rm ADV}, w_{\rm CAL}, w_{\rm OOD}]$ and $\mathbf{s} = [s_{\rm ID}, s_{\rm DS}, s_{\rm ADV}, s_{\rm CAL}, s_{\rm OOD}]$, then
\begin{equation}
    s_{\rm HR} = \mathbf{w}^\top \mathbf{s} \text{.}
\end{equation}
The application domain should dictate the relative importance of each metric, however in all of our analyses, we set the weights equal to each other. We note that by averaging the scores together, we lose fine-grained detail about the individual performance characteristics of a model in order to obtain a scalar value for reliability. This value is therefore not comprehensive, but is another tool by which models can be evaluated.

\subsection{Selection of ML Algorithms and Training Approach}
Our analysis relies on having many models trained using various ML algorithms on each of the datasets. Another benefit of using the WILDS benchmarks is that pre-trained models are publicly available, and we make use of these models. However, the algorithms used to train these models do not cover the full spectrum of approaches for reliable ML so we must also train our own models. 

\paragraph{WILDS Pre-Trained Models} In the original WILDS benchmark work~\citep{koh2021wilds}, the authors focused on approaches that promoted distribution-shift robustness by promoting invariance in the model outputs or optimizing for worst-case performance. The following approaches were explored: Invariant Risk Minimization (IRM)~\citep{arjovsky2019invariant}, Correlation Alignment (CORAL)~\citep{sun2016deep}, and Distributionally Robust Optimization (groupDRO)~\citep{Sagawa2020Distributionally}. In the second version of the benchmark~\citep{sagawa2022extending}, the focus was on the use of unlabeled training data to improve distribution-shift robustness. Approaches that promote invariance using unlabeled data were explored such as Domain-Adversarial Training of Neural Networks (DANN)~\citep{ganin2016domain}, Adaptive Feature Norm (AFN)~\citep{xu2019larger}, and again CORAL, but with additional domains provided by the unlabeled data. Approaches that used self-training were also explored such as PseudoLabels~\citep{lee2013pseudo}, Noisy Student~\citep{xie2020self}, and FixMatch~\citep{sohn2020fixmatch}. Additionally, one contrastive approach, Swapping Assignments between multiple Views (SwAV)~\citep{caron2020unsupervised}, was also used. Baseline models using empirical risk minimization (ERM) were trained for each version of the benchmark, and on the second version, ERM with RandAugment~\citep{cubuk2020randaugment} was also tried. We use the non-augmented ERM models as our baselines when evaluating the effect of various reliability-promoting ML approaches.

For each algorithm, hyperparameters were selected by choosing the best configuration from 10 random hyperparameter samples. For this set of hyperparameters, between 3 and 10 models were trained, each with different random seeds. We ignore the models trained during hyperparameter tuning and use only the hyperparameter-optimized models in our analysis. All \num{234} pre-trained models we used are publicly available.\footnote{\url{https://worksheets.codalab.org/worksheets/0x52cea64d1d3f4fa89de326b4e31aa50a}}

\paragraph{Selection of Other Reliable ML Algorithms} To decide which additional approaches we explore, we used the following criteria:
\begin{enumerate}
    \item The approach should show at or near SOTA performance on a large-scale benchmark (such as ImageNet) for at least one reliability criterion we are evaluating.
    \item The approach should be generalizable to all datasets we consider, therefore approaches that require domain expertise are excluded.
    \item The approach should be feasible to implement given the time and resources available. We therefore favor approaches that have open-source availability and lower implementation overhead. 
\end{enumerate}

Using these criteria, we selected the following set of approaches to evaluate (additional details in \cref{app:experimental_setup} and code provided\footnote{\url{https://github.com/sisl/holistic-reliability-evaluation}}):
\begin{itemize}
    \item \textbf{Dataset augmentation}: We experiment with two augmentation techniques. AugMix~\citep{hendrycks2020augmix} layers a series of simple image augmentation operations in concert with a consistency loss and has been shown to halve the error rate of corrupted CIFAR-10. The second technique, RandAugment~\citep{cubuk2020randaugment}, randomly selects from a pool of augmentation transformations to apply sequentially to an image and has been shown to achieve SOTA performance on benchmark datasets like CIFAR-10, SVHN, and ImageNet.
    \item \textbf{Adversarial data augmentation}: Following the approach of \citet{madry2018towards}, we use Projected Gradient Descent (PGD) to adversarially perturb each input during training. We use \num{10} iterations of PGD for each batch causing the training time to increase by a factor of about 10. Despite it being several years old, this approach still seems to be among the most effective for adversarial robustness~\citep{xuAdversarialAttacksDefenses2020}. 
    \item \textbf{Fine-tuning from pre-trained models}: Some large improvements in ID performance and DS robustness have been observed when fine-tuning pre-trained models~\citep{wortsman2022robust,kumar2022fine,radford2021learning}. We therefore explore this methodology by using weights that have been pre-trained using a variety of algorithms and across a variety of architectures. 
    \item \textbf{Model ensembling}: Ensembling of models has been shown to improve ID performance, DS robustness, and calibration~\citep{lakshminarayanan2017simple,wortsman2022model}. We investigate the composability of models trained as part of the WILDS benchmark. We combine the outputs  of up to \num{5} models using scalar weights that are optimized on the ID validation set. 
    \item \textbf{Temperature scaling}: Temperature scaling is a simple yet effective approach for the post-hoc calibration of deep neural networks and outperforms many other approaches. We compare reliability metrics pre- and post-temperature scaling. 
\end{itemize}
To avoid the costly overhead of tuning hyperparameters for each training configuration, instead we randomized the hyperparameters for each experiment.  We included all of the trained models in our analysis, even though some would have sub-optimal  hyperparameter configurations. In our analysis, we therefore focus on the best-performing set of models. In total, we trained an additional \num{287} models to bring the total models analyzed above \num{500}. We evaluated each of these models with and without temperature scaling on both validation and test sets. 

\subsection{Limitations}
Prior to discussing the results of our experiments, we want to outline the key limitations of our experimental approach. 

\paragraph{Incomplete Assessment of Reliability} Our workflow focuses on only five criteria: in-distribution performance, distribution-shift robustness, adversarial robustness, calibration, and out-of-distribution detection. While these metrics cover important aspects of reliability, they may not capture the full range of challenges and limitations associated with evaluating machine learning systems. Other relevant factors such as fairness/bias, interpretability, and privacy are not adequately addressed by the selected metrics, however could be included if they are measured quantitatively. Additionally, safety and security are features of the complete system in the socio-technical context within which it operates~\citep{leveson2016engineering}. Without that context, it is difficult to draw general conclusions around the safety of AI systems. 

\paragraph{Choice of Quantitative Metrics} For each of the reliability criteria we do analyze, we measure a single number to indicate performance, but often there are multiple relevant metrics. For example, when measuring performance, we focus on top-1 accuracy but other metrics such as top-$k$ accuracy, confusion, or subgroup accuracy may all be relevant depending on the task. Measuring adversarial robustness assumes a specific set of attack strategies, and uncertainty quantification could be measured by metrics such as integrity risk~\citep{hansen2004waas} rather than calibration. For OOD detection, we assume a specific set of input faults, which may not reflect the types of inputs we would wish to flag during model deployment. Additionally, we selected $\mathit{AUROC}$ to quantify the capability of the OOD detector, but other metrics such as a false positive rate at a fixed true positive rate may also be relevant depending on the specific context of the task.

\paragraph{Lack of Target Thresholds} To assess the reliability of ML models it would be helpful to have target values of our evaluation criteria in mind to quantify how close our models are to being sufficiently reliable. However, acceptable thresholds for these parameters are context dependent. For example, in aviation, failure rate requirements can be as low as \num{e-9} per hour~\citep{do178}, while in some medical settings, we tolerate much higher failure rates when the alternative is also risky. We therefore do not include such thresholds in our analysis. 

\paragraph{Limited Exploration of Algorithms and Hyperparameters} Due to the high computational cost of training many different ML models across several datasets, we are unable to explore every algorithm for building reliable ML models and to exhaustively search the space of hyperparameters (neither did the researchers who trained the WILDS pre-trained models we analyze). Therefore, when we observe no impact (or a negative impact) of an algorithm, it does not constitute proof that the approach is ineffective and there might be a configuration of hyperparameters that would improve the impact.

\paragraph{Limited Real-World Tasks} While the WILDS benchmark tasks offer a great framework to study real-world distribution shifts, there are many more real-world domains of high consequence and the generalizability of our findings should be explored on each new domain. \\

While the proposed workflow and evaluation metrics provide a valuable framework for assessing the reliability of machine learning systems, it is important to recognize these limitations. Consideration of additional factors and context-specific requirements are necessary to develop a more comprehensive and robust evaluation approach.

\section{Results}

In this section, we present the results of our experiments. We start by exploring the effect of different ML interventions for improving model reliability and then consider the relationship between reliability metrics. When possible, we show the results from all three datasets simultaneously. When assessing individual algorithms, however, we present the results of our experiments on the iWildCam dataset, then present the results for the remaining datasets in \cref{app:additional_results}. Unless otherwise mentioned, the conclusions we draw from the iWildCam results are consistent with the results on the other datasets.

\subsection{Data Augmentation}

\begin{figure}[ht]
    \centering
    \includegraphics[width=\textwidth]{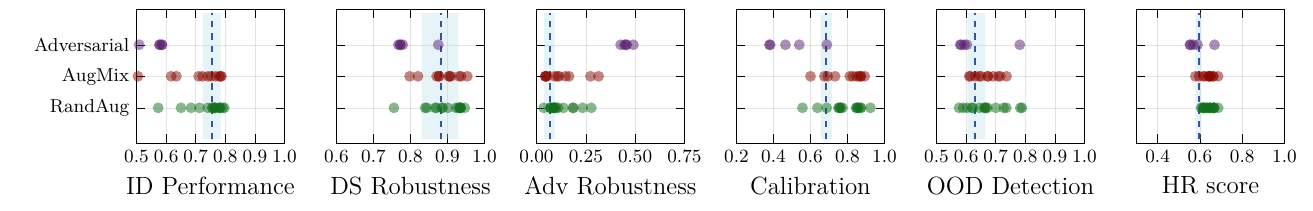}
    \caption{Effect of various data augmentation strategies for iWildCam. Each circle is a single model. The blue band shows the range of the ERM baseline models with the dashed line giving the average.}
    \label{fig:dataset_augmentation_iwildcam}
\end{figure}

We experimented with two common forms of data augmentation: \emph{RandAug}~\citep{cubuk2020randaugment} and \emph{Augmix}~\citep{hendrycks2020augmix}, as well as \emph{PGD}-based adversarial training~\citep{madry2018towards}. The results for iWildCam are shown in \cref{fig:dataset_augmentation_iwildcam} and the results for the remaining datasets are in \cref{app:data_augmentation}. Adversarial training primarily seemed to affect in-distribution performance (where it reduced slightly) and adversarial robustness where it tended to increase it significantly, especially on iWildCam and FMoW, which have a larger number of classes and are therefore more susceptible to adversarial attack. While adversarial training reduced calibration and distribution-shift robustness in the worst case for iWildCam, it did not have a consistent effect for the other datasets, so we draw no general conclusions. Notably, the highest performing RandAug and Augmix models improved across all of the metrics (increasing the HR score by about \num{0.08}), with the largest gains coming in calibration, which is consistent for the other datasets. 

\subsection{Model Ensembling}
\begin{figure}[ht]
    \centering
    \includegraphics[width=\textwidth]{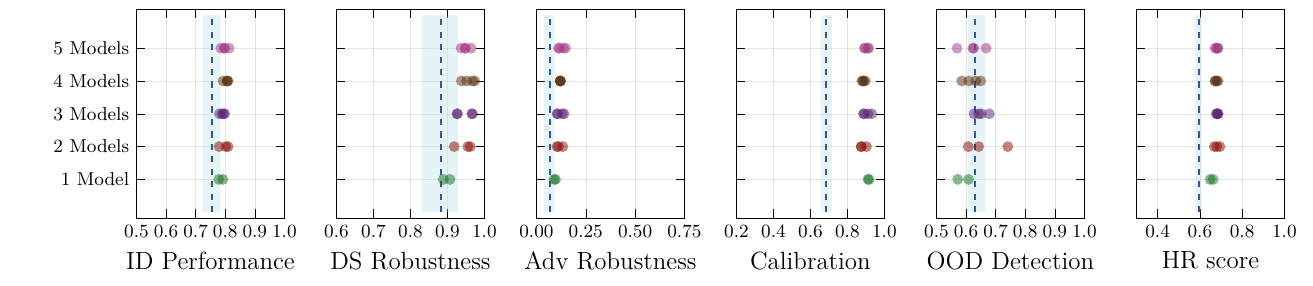}
    \caption{Ensembles of models for iWildCam.}
    \label{fig:ensembles_wilds_iwildcam}
\end{figure}

To enhance the reliability of non-ML-based systems, it is common to compose or ensemble redundant components to reduce errors and the likelihood of faults. In this experiment, we investigate the effect of ensembling ML-based components. We randomly choose different numbers of models and combine their logits with weights optimized for ID performance on a validation set. Each point in \cref{fig:ensembles_wilds_iwildcam} represents the best ensemble (in terms of ID validation performance) from a selection of \num{50} random ensembles of the same size. Different groups of \num{50} were used to get the different points in \cref{fig:ensembles_wilds_iwildcam}. For iWildCam, ensembling improves nearly all of the reliability metrics (and therefore HR score) except for OOD detection, which remains largely unaffected. We note, however, that uncertainty estimates from ensembles can be used as a strategy for OOD detection~\citep{lakshminarayanan2017simple}, but we do not apply that technique here. The gains from ensembling appear to diminish as the number of models increases i.e. using five models is not much better than four in all of the datasets. This finding is consistent with prior work on ensembles~\citep{gontijo2021no}.

\subsection{Fine-Tuning Pre-Trained Models}
For these experiments, we assess the effect of fine-tuning models that have been pre-trained in a variety of ways. The fine-tuning procedure is specified in more detail in \cref{app:fine_tuning_experimental_setup}. We first explore the type of pre-training, then discuss model architecture, and lastly experiment with the size of pre-trained transformers.

\begin{figure}[ht]
    \centering
    \includegraphics[width=\textwidth]{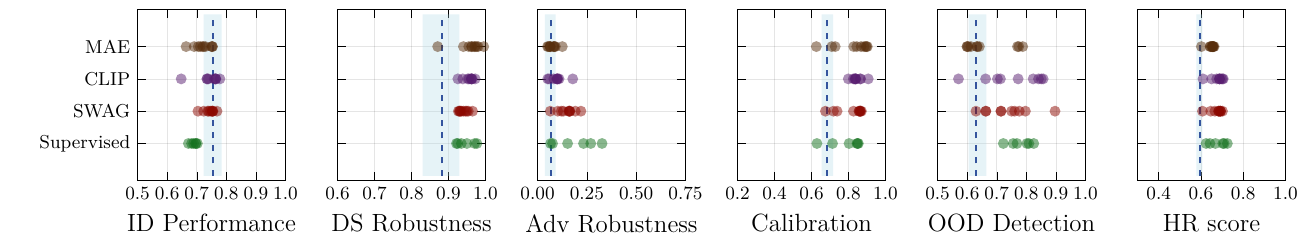}
    \caption{Effect of pre-training algorithm when fine-tuning on iWildCam.}
    \label{fig:pretraining_iwildcam}
\end{figure}

\paragraph{Pre-Training Algorithm} Our goal is to see if the type of pre-training is important for the final performance of the pre-trained model. We compare four pre-training algorithms: \emph{Supervised} is pre-trained using supervised learning on ImageNet-1k, SWAG is Supervised Weakly from hashtAGs~\citep{singh2022revisiting}, \emph{CLIP} is Contrastive Language-Image Pre-training~\citep{radford2021learning}, and \emph{MAE} is for masked-autoencoder~\citep{he2022masked}, a generative pre-training technique. The results for iWildCam are shown in \cref{fig:pretraining_iwildcam}.  All of the pre-training techniques seem to substantially improve  DS robustness, calibration, and OOD detection, leading to higher overall HR scores ($>0.1$ improvement). MAE pre-training seems to most reliably lead to the highest DS robustness. Given the focus on unsupervised pretraining as a method for improving robustness~\citep{wortsman2022robust} it is notable that even supervised pretraining provided significant reliability benefits across all datasets. 

\paragraph{Model Architecture} To explore the influence of model architecture when fine-tuning models, we use a variety of architectures, each pre-trained via supervised learning on ImageNet-1k. We compare the Vision Transformer (\emph{ViT})~\citep{dosovitskiy2021an},\emph{Swin Transformer}~\citep{liu2021swin},  \emph{MaxViT}~\citep{tu2022maxvit}, \emph{ConvNeXt}~\citep{liu2022convnet} and \emph{EfficientNet}~\citep{tan2019efficientnet} architectures. In the context of supervised pre-training, the Swin Transformer, EfficientNet and ConvNeXt architectures had the best ID performance, while all the transformers and ConVeXt architecture had superior DS robustness, calibration and OOD detection. On the other datasets, however, only the ViT and ConvNext architectures showed significant DS robustness improvements.

\begin{figure}[ht]
    \centering
    \includegraphics[width=\textwidth]{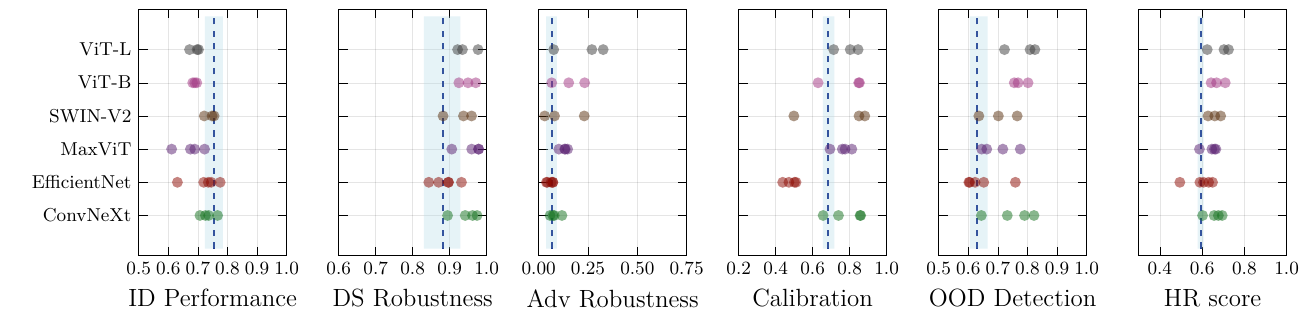}
    \caption{Effect of model architecture when fine-tuning on iWildCam.}
    \label{fig:model_type_iwildcam}
\end{figure}

\begin{figure}[ht]
    \centering
    \includegraphics[width=\textwidth]{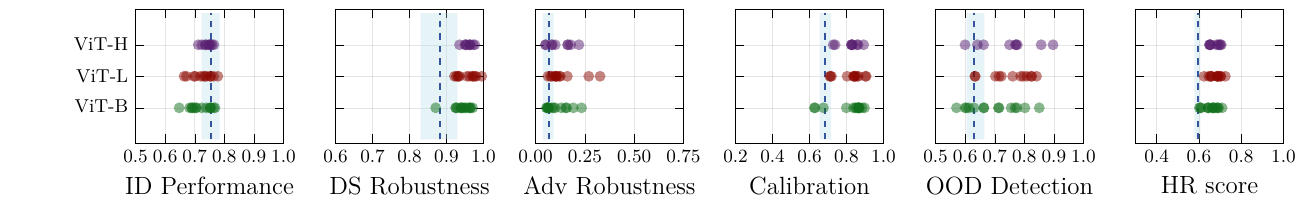}
    \caption{Effect of transformer size when fine-tuning on iWildCam.}
    \label{fig:model_size_iwildcam}
\end{figure}

\paragraph{Model Size} Lastly, for a fixed model architecture (ViT) we explore the effect of model size by comparing the base \emph{ViT-B}, large \emph{ViT-L} and huge \emph{ViT-H} model versions. We show the result for iWildCam in \cref{fig:model_size_iwildcam}. The larger models show slight improvements over smaller models in almost all categories. When considered with the other datasets, the largest transformer seemed to be less sensitive to hyperparameters, giving consistently good performance.

\subsection{Sub-Group Aware Training}
\begin{figure}[ht]
    \centering
    \includegraphics[width=\textwidth]{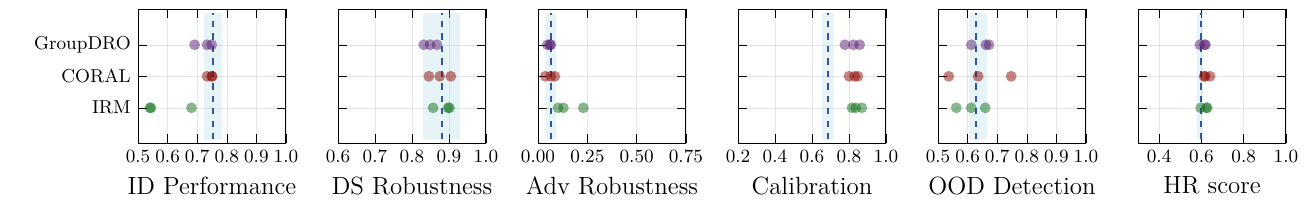}
    \caption{Effect of including sub-group information for iWildCam.}
    \label{fig:invariant_loss_iwildcam}
\end{figure}

Prior work in improving distribution shift robustness has explored the use of group invariance in the features output by the model. The WILDS benchmark trained models with GroupDRO~\citep{Sagawa2020Distributionally}, correlation alignment (CORAL)~\citep{sun2016return,sun2016deep} and invariant risk minimization (\emph{IRM})~\citep{arjovsky2019invariant}, which we analyze further by measuring all their reliability metrics. The results for iWildCam are shown in \cref{fig:invariant_loss_iwildcam}. While these algorithms do not improve the DS robustness, they do have a small positive impact on calibration for iWildCam. When considering the other datasets, however, these approaches have limited impact on any of the reliability metrics including HR score.

\subsection{Unlabeled Training Data}
\begin{figure}[ht]
    \centering
    \includegraphics[width=\textwidth]{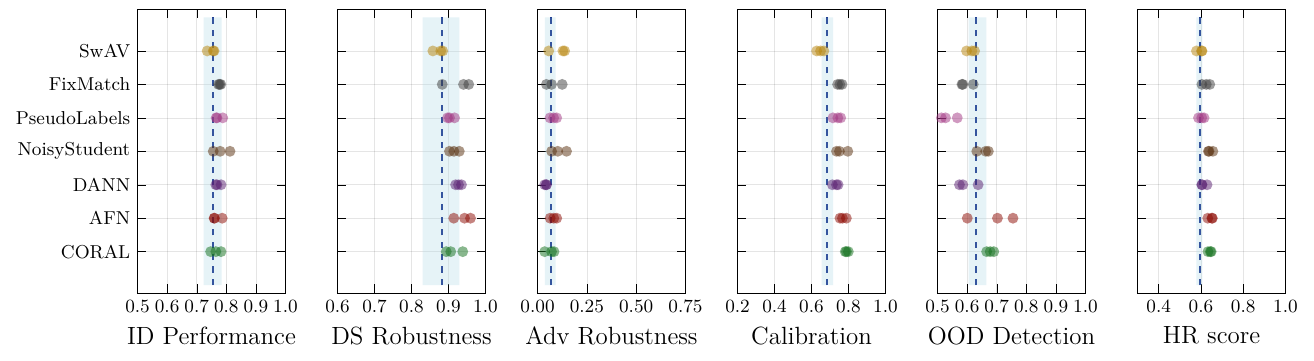}
    \caption{Effect of using unlabeled training data for iWildCam.}
    \label{fig:unlabeled_data_iwildcam}
\end{figure}

For many tasks, unlabeled data is readily available and can be used in a variety of ways to improve model performance. \citet{sagawa2022extending} investigated the use of unlabeled data for improving robustness to distribution shift. They investigated techniques that labeled the unlabeled data (\emph{PseudoLabels}~\citep{lee2013pseudo}, \emph{Noisy Student}~\citep{xie2020self}, \emph{FixMatch}~\citep{sohn2020fixmatch}), that used sub-group aware training on the unlabeled data (Domain-Adversarial Training of Neural Networks (\emph{DANN})~\citep{ganin2016domain}, Adaptive Feature Norm (\emph{AFN})~\citep{xu2019larger}, and Correlation alignment (CORAL)~\citep{sun2016return,sun2016deep}), and a contrastive learning technique using the unlabeled dataset (Swapping Assignments between multiple Views (\emph{SwAV})~\citep{caron2020unsupervised}). We use the models trained by \citet{sagawa2022extending} and see how the unlabeled data affected all reliability metrics (results for iWildCam shown in \cref{fig:unlabeled_data_iwildcam}). For iWildCam we can see that many of the approaches lead to a small improvement in DS robustness as well as calibration and leave the other metrics largely unchanged. There was a larger improvement to DS robustness for the FMoW dataset and larger improvements on calibration and OOD detection for the Camelyon17 dataset, leading to a significant improvement in HR score. For each dataset, the algoriths that performed best for each metric differed.   

\subsection{Temperature Scaling}

\begin{figure}[ht]
    \centering
    \includegraphics[width=\textwidth]{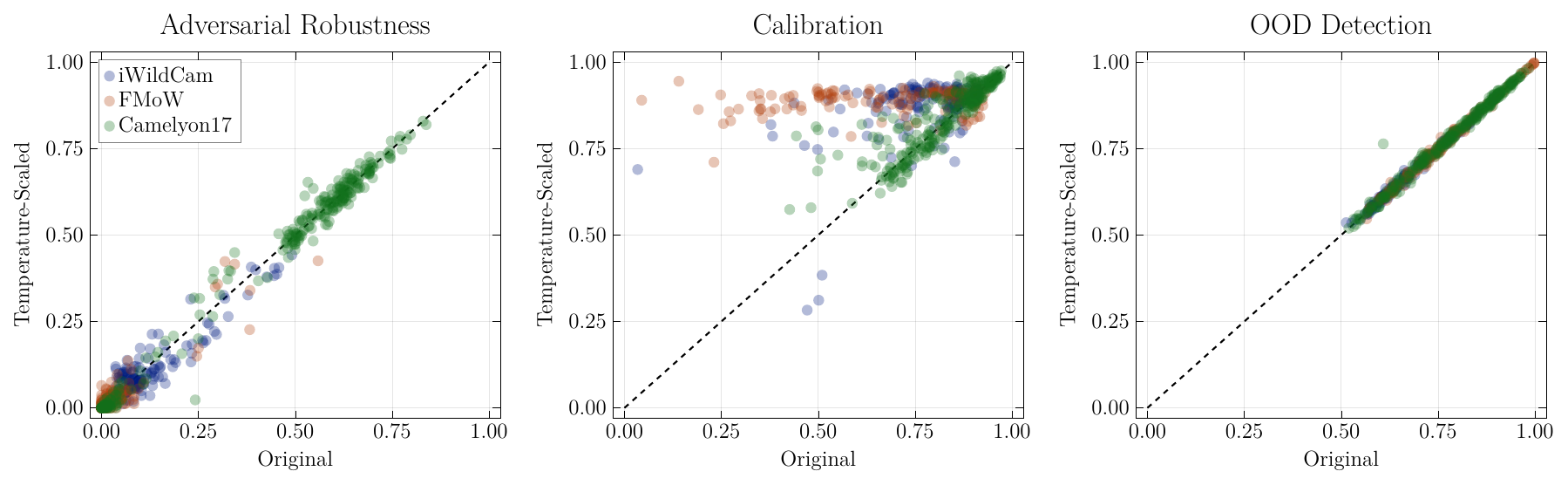}
    \caption{Effect of temperature scaling on model calibration, adversarial robustness and OOD detection for all datasets.}
    \label{fig:temperature_scaling}
\end{figure}

Temperature scaling has been shown to be a simple and effective technique for post-hoc calibration of deep neural networks. For each model we trained, we computed the reliability metrics both before and after temperature scaling. The temperature parameter was optimized by minimizing the negative log-likelihood of the in-distribution validation set. Since temperature scaling does not change the model prediction, it has no effect on ID performance nor DS robustness so we exclude those metrics in the results shown in \cref{fig:temperature_scaling}. First, we find that there is no strong effect on adversarial robustness nor OOD detection, but calibration is generally improved. For Camelyon17, temperature scaling had less of an impact than in the other datasets, making the calibration of the non-temperature-scaled models more important. Temperature scaling therefore increases HR score by improving calibration without affecting other metrics. 

\subsection{Distribution of Scores and Summary of Improvements}
In order to summarize the effects of our experiments, we report the performance of approaches grouped into conceptual categories. Similar to the breakdown of the previous subsections, we categorize the models into seven approaches: \emph{Baselines}, which includes vanilla empirical risk minimization, \emph{Data Augmentation}, which includes RandAug and AugMix, \emph{Adversarial Training}, which includes just the PGD-based adversarial training, \emph{Ensembles}, which includes all the ensembles, \emph{Fine-Tuned}, which includes all of the fine-tuned models, \emph{Loss Function}, which includes the sub-group aware training approaches, and \emph{Unlabeled Data}, which includes all of the approaches that used unlabeled training data. 

\begin{figure}[ht]
    \centering
    \includegraphics[width=\textwidth]{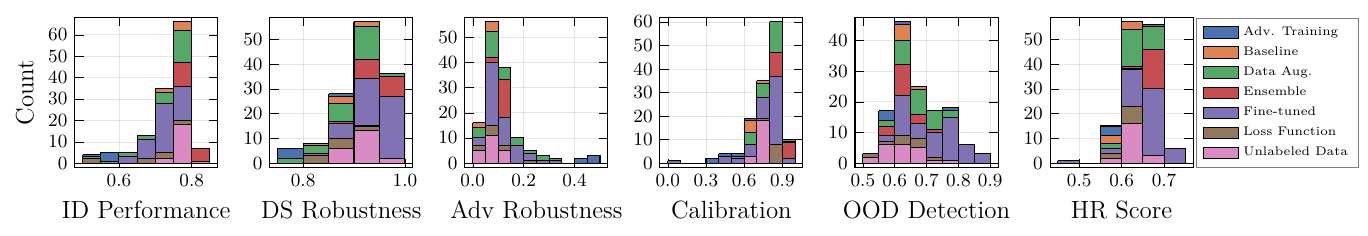}
    \caption{Distribution of scores for iWildCam for various groups of models.}
    \label{fig:distribution_of_scores_iwildcam}
\end{figure}

\cref{fig:distribution_of_scores_iwildcam} Provide a comparison of the distribution of scores across groups. We plot a group-stacked-histogram of each reliability metric for iWildCam for the seven groups. The results from the other datasets are included in \cref{app:additional_results}. The aggregate influence of each type of approach can then be compared. Ensembling has the biggest improvement on ID performance and calibration, while fine-tuning has the biggest effect on DS robustness, OOD detection and HR score. Only the adversarial training was able to consistently achieve higher levels of adversarial robustness.

\begin{figure}[ht]
    \centering
    \includegraphics[width=\textwidth]{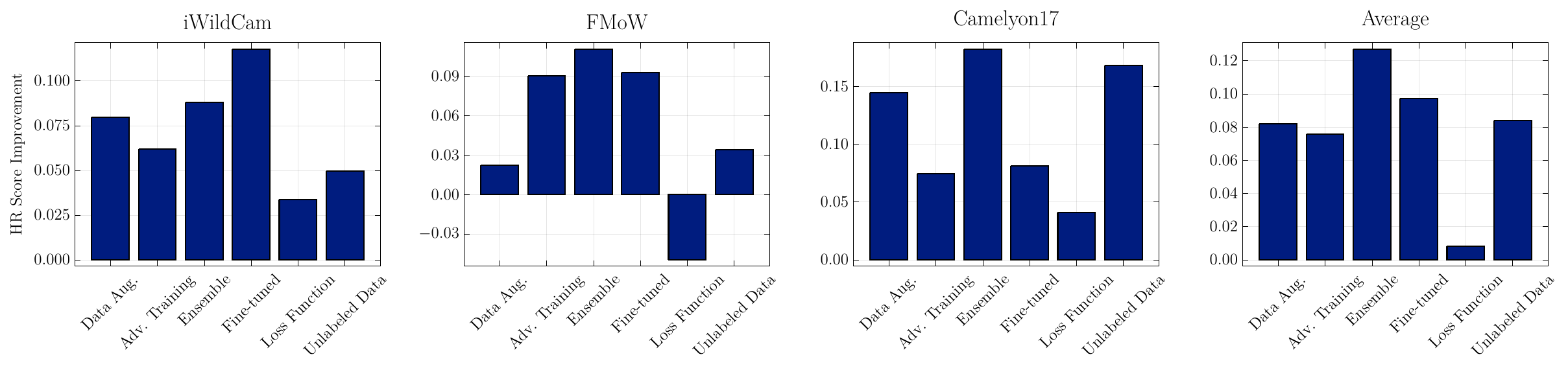}
    \caption{HR score improvement for highest HR score model in each group compared to the highest HR score baseline model.}
    \label{fig:hr_score_max_improvements}
\end{figure}

 \Cref{fig:hr_score_max_improvements} summarizes the quantitative influence of each group of approaches on the HR score. We plot the difference between the best HR score in each group and the best HR score for the baselines across the three datasets (as well as the the average of this value). We see improvements from all groups of approaches except those that use auxiliary loss functions. We note that the lower effect of fine-tuning on the Camelyon17 dataset is primarily due to low values of adversarial robustness for the fine-tuned models. 

\subsection{Metric Comparisons}

\begin{figure}[ht]
    \centering
    \includegraphics[width=\textwidth]{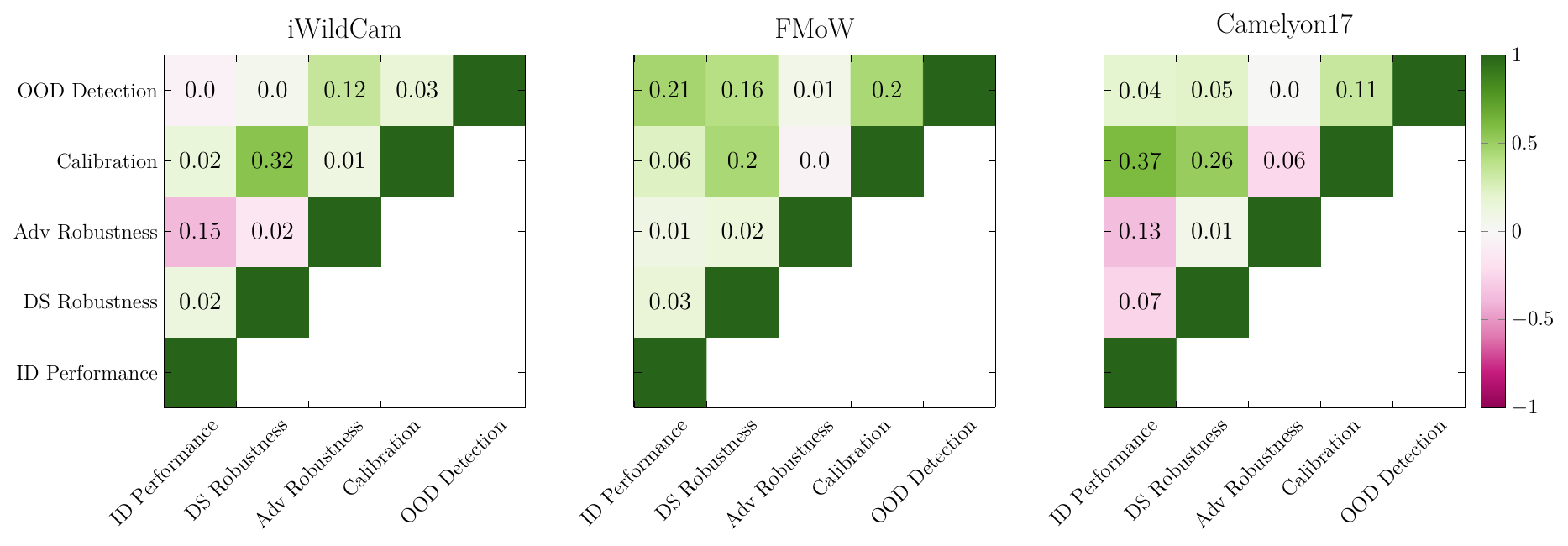}
    \caption{Correlations between reliability metrics for three datasets. Color indicates Pearson correlation and value indicates $R^2$.}
    \label{fig:metric_correlations}
\end{figure}

Motivated by prior work that observes correlations between a variety of metrics such as in-distribution accuracy vs. distribution-shifted accuracy~\citep{miller2021accuracy}, adversarial accuracy vs. in-distribution accuracy~\citep{schmidtAdversariallyRobustGeneralization2018,ilyasAdversarialExamplesAre2019,tsipras2018robustness}, we explore the possibility of intrinsic correlations between all five reliability metrics. 
To remove the correlations induced by using different machine learning algorithms we first group the approaches as before into seven groups. Then, for each group, we subtract off the mean for each metric, making it centered around \num{0}. We then combine the zero-centered data of each group into one dataset from which we compute the Pearson correlation and $R^2$ values for each pair of metrics. The results are shown in \cref{fig:metric_correlations} where the color corresponds to correlation (and can be negative or positive) while the number corresponds to $R^2$, which is strictly positive. Values of \num{1} along the diagonal are where each metric is compared to itself, and we only show one set of off-diagonal values as these correlation metrics are symmetric.

For almost all pairs of metrics there does not appear to exist a strong relationship between them, with most of the $R^2$ values being less than \num{0.2}. This finding is consistent with the observation that deep neural networks are underspecified~\citep{d2020underspecification} so designing to a particular metric (such as in-distribution performance) does not constrain other metrics (hence the low correlation). There are, however, a few exceptions. There is a positive correlation between DS robustness and calibration for all three datasets. This effect seems to primarily come from the fine-tuned models and is discussed further in \cref{app:metric_correlations}. There is a weak negative correlation between ID performance and adversarial robustness (a relationship previously observed~\citep{schmidtAdversariallyRobustGeneralization2018}). There are also dataset-specific relationships. For FMoW, there is a correlation with OOD detection and several other metrics, while ID performance and calibration are related for Camelyon17. We note that these correlations (and lack thereof) may be specific to the range of scores achieved by the models in this study. Additional relationships may exist for lower or higher performing models.

\section{Discussion}
\label{sec:discussion}
In this section, we first distill our investigation into the reliability of machine learning systems, highlighting three categories of findings: the pervasive issue of underspecification, the effects of targeted algorithmic improvements, and the broader reliability enhancements achieved by certain techniques. We then propose some research questions that emerged out of our analysis and lastly we conclude. 

\subsection{Summary of Key Findings}
Here we summarize the key findings from our holistic assessment of the reliability of machine learning systems. 

\paragraph{Underspecification} In accordance with prior work~\citep{gulrajani2021in,d2020underspecification}, we find that deep neural networks are underspecified (selecting for one metric does not constrain the others) across all reliability metrics we measure. We found almost no intrinsic correlation between any of the reliability metrics when measured on over \num{500} models trained in diverse ways across three real-world datasets. This conclusion has several key implications. Firstly, it underscores the critical need for comprehensive evaluation methodologies in the development of machine learning algorithms, of which the HR score is a small step towards. Not only should individual reliability metrics be assessed and optimized, but understanding their interdependencies and the potential trade-offs is equally essential. Secondly, it highlights the importance of model selection prior to deployment. Any reliability metric that is not measured and selected could take on a wide range of values, potentially hurting overall system performance. Lastly, the lack of connections between these metrics means there can be substantial value in developing techniques that improve one of these reliability metrics in isolation (like temperature scaling does for calibration). These bespoke methods could then be combined to construct high-reliability models in a more modular fashion.

\paragraph{Isolated Improvements} Some algorithms we investigated lead to improvements on just one of the metrics and sometimes at the cost of others. Adversarial training was the only approach to make substantial and consistent improvements in adversarial robustness, but in the case of iWildCam it always came at the cost of in-distribution performance and distribution shift robustness. Temperature scaling nearly always improved calibration (though it was the least effective on the binary classification task of Camelyon17), and leaves the other metrics unchanged. Learning from unlabeled data lead to modest improvements in DS robustness and sometimes calibration but the improvements varied significantly across datasets and connection to specific algorithms was tenuous. Sub-group aware training methods failed to consistently improve DS robustness, the metric they were designed for, but did seem to have a positive effect on calibration. 

\paragraph{Comprehensive Reliability Improvements} Other techniques, however, had substantial improvements to reliability across almost all metrics. Ensembling models improved all metrics except OOD detection leading to an average increase of \num{0.12} in HR score across datasets. There was an average increase of about \num{0.08} in HR score for the best models under the two non-adversarial data augmentations where the improvements came in every metric except in-distribution performance. Fine-tuning large pre-trained models had large improvements for all three datasets with an average increase of HR score by about \num{0.08}.

\subsection{Future Research Directions}
Our study has highlighted several important questions and research gaps that warrant further exploration.

\paragraph{Improved Understanding of Pre-Trained Representations} Pre-trained models offer robust performance and calibration, along with OOD detection capabilities, but the reasons for their effectiveness remain unclear. Preliminary work, such as by \citet{caron2021emerging}, suggests that features in pre-trained ViTs contain semantic segmentation information, contrasting with their supervised counterparts. Additional research is needed to understand these representations and how they arise. 

\paragraph{Refining Fine-Tuning Dynamics} To leverage the full potential of fine-tuning, we must explore its multidimensional design space. This includes understanding which architectures are best suited for fine-tuning, the ideal pairing of pre-training datasets with architectures and downstream tasks, and the optimization strategies that yield the best results. Moreover, exploring whether fine-tuning can enhance adversarial robustness is of interest, as suggested by \citet{jeddiSimpleFinetuningAll2020}.

\paragraph{Implications of Pre-Trained Models as Backbones} If pre-trained models will form the foundation of highly reliable machine learning systems, then it will be crucial to ensure their widespread availability and to safeguard them against bias and security vulnerabilities. Much work regarding the widespread, fair, and secure deployment of large pre-trained models will fall into the domain of policy and governance~\citep{zwetsloot2018beyond, birhane2022values, benderdangerparrots}. From a technical standpoint, a variety of reliability metrics and vulnerabilities should be investigated. Recent work~\citep{2022arXiv221109110L} has performed a holistic evaluation of language models and covers many of the same evaluation metrics, but focused on natural language tasks. ML systems are also susceptible to backdoor trojan attacks where models are manipulated by an adversary to have different behavior when a particular set of circumstances occurs~\citep{gu2019badnets}. Work in this area, such as the TrojAI Software Framework from \citet{Karra2020TheTS}, allows researchers to rapidly and comprehensively test new trojan detection methods. Cybersecurity risks are another major security vulnerability when deploying large pre-trained models; work in this area, for example, could build upon efforts to leverage ML to detect intruders of these systems~\citep{networkintrusion}. Beyond these concerns, a challenge lies in devising equitable mechanisms for shouldering the developmental costs of these foundational models, thus broadening their benefits.

\paragraph{Effective Use of Task-Specific Unlabeled Data} Despite notable reliability improvements achieved by training on extensive corpora of unlabeled data, the use of task-specific unlabeled data was shown to be less successful. This discrepancy might stem from the chosen algorithmic approaches or the relative size disparity between these datasets and those used in models like CLIP. Further investigation into effective utilization of task-specific unlabeled data is needed, with a promising starting point in approaches that achieve SOTA performance on ImageNet without much extra data such as Data-efficient Image Transformers~\citep{touvron2021training} and Masked Autoencoders~\citep{he2022masked}.

\paragraph{Establishing Challenge Problems to Determine Requirements} Comparing machine learning models' reliability across various metrics is straightforward; the challenge lies in determining when they are \emph{reliable enough}. Understanding the context of these systems is vital for gauging the required reliability properties. A productive endeavor could be to establish real-world challenge problems (not just benchmarks), where machine learning models form crucial but not stand-alone components. Researchers would be free to design systems around these components to help ensure reliability, and from these designs may emerge target requirements that individual components need to achieve. Having such requirements could help drive research and inform regulation.

\subsection{Conclusion} This study conducted a holistic assessment of machine learning systems, focusing on their reliability across diverse metrics. Findings confirmed the phenomenon of underspecification in deep neural networks, emphasizing the necessity of comprehensive evaluation methodologies and careful model selection prior to deployment. While some algorithmic approaches showed improvements on specific metrics, often at the expense of others, techniques such as ensembling and fine-tuning large pre-trained models yielded considerable improvements in holistic reliability. As AI continues to be increasingly deployed in high-stakes applications, we hope that our findings guide future research to develop more reliable and robust systems.

\section{Acknowledgements}
This work was supported by the Foundational Research Grants program at Georgetown University's Center for Security and Emerging Technology. We would like to thank Helen Toner, Dewey Murdick, Drew Lohn, Andrea Guerrero, Esen Yel, Robert Moss, Josh Ott, Mansur Arief, and Sydney Katz for their valuable feedback and insightful discussion. 

\begingroup
\sloppy
\printbibliography
\endgroup

\appendix

\section{Additional Experimental Setup Details}
\label{app:experimental_setup}

Here we describe various implementation details of the algorithmic approaches we applied when training models for the experiments. All code is publically available.\footnote{\url{https://github.com/ancorso/holistic-reliability-evaluation}} We conducted our experiments on NVIDIA A100 and V100 GPUs, with the V100s dictating the maximum batch size due to their lower memory of \num{12}GB.

\subsection{Evaluation Metric Details}

Below we include additional details regarding the computation of the five reliability metrics. 
\begin{itemize}
    \item \textbf{ID Performance}. To compute ID performance we used \num{1024} samples from the in-distribution test set provided by the WILDS benchmark. For the Camelyon17 dataset, we scaled the ID performance metric to linearly map values from ($0.7$, $1.0$) to ($0.0$, $1.0$) since all models achieved high performance ($>0.7$).
    \item \textbf{DS Robustness}. To compute the DS robustness, we evaluated each model on three distribution shifts, the real-world ``val'' and ``test'' shifts provided by the WILDS Benchmark and a synthetic corruption shift inspired by ImageNet-C~\citep{hendrycks2018benchmarking}. For each shift we compute the DS performance and divide it by the ID performance and average the results. 
    \item \textbf{Adversarial Robustness}. To compute adversarial robustness, we use the AutoAttack package\footnote{\url{https://github.com/fra31/auto-attack}} to adversarially perturb \num{128} images and take the ratio of the adversarial performance to ID performance.
    \item \textbf{Calibration}. We use the built-in calibration error function provided by torchmetrics\footnote{\url{https://torchmetrics.readthedocs.io/en/stable/classification/calibration_error.html}} to compute expected calibration error for the ID predictions and the DS predictions and average the resutls. 
    \item \textbf{OOD Detection} For each task we used data from the remaining tasks as well as Gaussian noise and RxRx1 data for OOD detection. For example, the OOD detection task for iWildCam consisted of distinguishing between iWildCam data and mixed data from Camelyon17, FMoW, RxRx1 and Gaussian noise. We use the implementation of OOD detection algorithms provided by the package pytorch-ood\footnote{\url{https://github.com/kkirchheim/pytorch-ood}}
    \item \textbf{HR Score}. We used even weights (\num{0.2}) to compute the HR score for all experiments. 
\end{itemize}

\subsection{Data Augmentation Experiments}
For these experiments we used a similar training procedure of the WILDS benchmark, while adding in different forms of data augmentation using the RandAug\footnote{\url{https://pytorch.org/vision/main/generated/torchvision.transforms.RandAugment.html}} and Augmix\footnote{\url{https://pytorch.org/vision/main/generated/torchvision.transforms.AugMix.html}} implementations built into torchvision. For iWildCam we used an image size of ($448$,$448$), a batch size of 24, and a Resnet-$50$ model. For FMoW we used an image size of ($224$,$224$), a batch size of 72, and a Densenet-$121$ model. For Camelyon17 we used an image size of ($96$,$96$), a batch size of 168, and a Densenet-$121$ model.
The learning rate was sampled from a log uniform distribution between \num{e-4} and \num{e-2}, label smoothing values were sampled from the set ($0.1$, $0.01$, $0.0$), and the optimizer options considered were ADAM, SGD, and ADAMW.

\subsection{Adversarial Data Augmentation Experiments}

For these experiments we used the same baseline models and input sizes as the data augmentation experiments. For iWildCam the batch size was set to \num{12}, for FMoW it was set to \num{64} and for Camelyon17 it was set to \num{256}. We used \num{10} iterations of the adversarial attack for each training batch, causing the training time to increase by a factor of about \num{10} when PGD is used. The key hyperparameters are the type of attack method, which we selected randomly between Projected Gradient Descent and the Fast Gradient Sign Method, and the strength of the attack, which we selected randomly from ($1/255$, $3/255$, $8/255$).

\subsection{Ensembling Experiments}
An ensemble was generated from a specified number of trained models by combining the logits of those models with a set of weights. The weights are optimized to minimize the loss on the ID validation set. Note that when only one model is present, this is the same process as temperature scaling (hence the improvement in calibration for just 1 ``ensemble'' model). To find the approximately best configuration of models to ensemble, we build \num{50} random ensembles (each with their own weights optimized on the validation set) and then we pick the one with the lowest validation loss. This best model is then evaluated across the other reliability metrics and included in our analysis. We repeated this experiment at least three times for each size of ensemble for each dataset. 

\subsection{Fine-Tuning Experiments}
\label{app:fine_tuning_experimental_setup}

For the fine-tuning experiments we obtained the pre-trained models from a variety of sources. The ViT, Swin Transformer, MaxViT, EfficientNet, and ConvNeXt models (with both supervised pre-training and SWAG pre-training) came from torchvision,\footnote{\url{https://pytorch.org/vision/stable/models.html}} the CLIP models came from open\_clip,\footnote{\url{ https://github.com/mlfoundations/open_clip}} and the MAE models came from the MAE repo.\footnote{\url{https://github.com/facebookresearch/mae}}

For these experiments, we resized images to ($224$, $224$ and used a batch size of \num{24}. We trained fro a maximum of \num{5} epochs, though we used validation accuracy to select the best model. When fine-tuning, we took the approach of freezing most upstream layers and then unfreezing some number of layers for for training. The number of layers we unfroze was a hyperparameter that could take on values of ($1$, $2$, $4$, $8$). The learning rate was sampled from a log uniform distribution between \num{e-4} and \num{e-2}, label smoothing values were sampled from the set ($0.1$, $0.01$, $0.0$), and the optimizer options considered were ADAM, SGD, and ADAMW.
To remove some variation due to hyperparameter selection,  we sampled a random configuration of hyperaparameters and trained all fine-tuned models on that configuration. We repeated with at least three different hyperparameter configurations for each datset.

\section{Additional Results}
\label{app:additional_results}
This section provides the results figures for the FMoW and Camelyon17 datasets that were not included in the main body of the paper. We provide some additional commentary on the results, though we draw no major conclusions outside of those summarized in \cref{sec:discussion}

\subsection{Dataset Augmentation}
\label{app:data_augmentation}

\begin{figure}[ht]
    \centering
    \includegraphics[width=\textwidth]{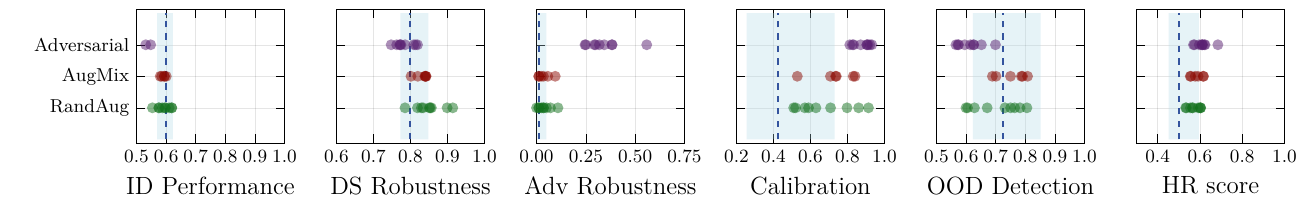}
    \caption{Effect of various data augmentation strategies for FMoW.}
    \label{fig:dataset_augmentation_fmow}
\end{figure}

\begin{figure}[ht]
    \centering
    \includegraphics[width=\textwidth]{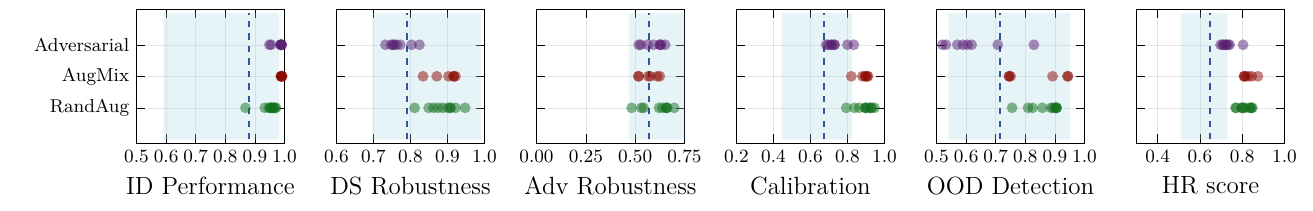}
    \caption{Effect of various data augmentation strategies for Camelyon17.}
    \label{fig:dataset_augmentation_camelyon17}
\end{figure}

Adversarial data augmentation drastically increases the adversarial robustness of FMoW (see \cref{fig:dataset_augmentation_fmow}) while having little effect on the Camelyon17 models (see \cref{fig:dataset_augmentation_camelyon17}), which already had higher levels of adversarial robustness. We speculate that adversarial robustness is harder to achieve on models with more classes as there are more decision boundaries that can be reached through perturbation. AugMix and RandAug led to strong performance improvements across most metrics for Camelyon17 with more modest improvements centered on calibration for FMoW.

\subsection{Model Ensembling}
\begin{figure}[ht]
    \centering
    \includegraphics[width=\textwidth]{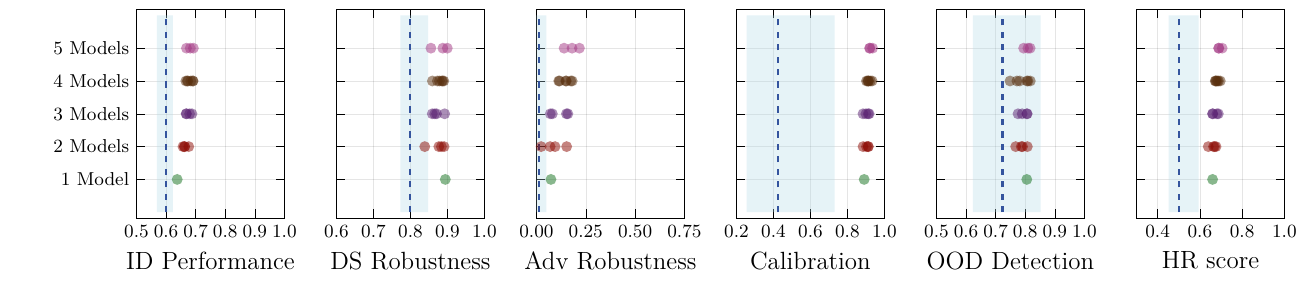}
    \caption{Ensembles of models for FMoW.}
    \label{fig:ensembles_wilds_fmow}
\end{figure}

\begin{figure}[ht]
    \centering
    \includegraphics[width=\textwidth]{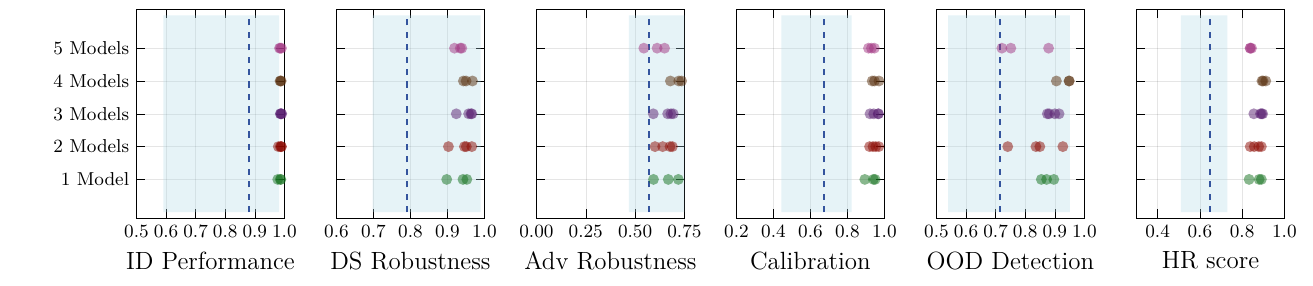}
    \caption{Ensembles of models for Camelyon17.}
    \label{fig:ensembles_wilds_camelyon17}
\end{figure}

Ensembling shows improvements across all metrics for both Camelyon17 and FMoW (see \cref{fig:ensembles_wilds_fmow} and \cref{fig:ensembles_wilds_camelyon17} respectively). The number of models where performance saturates seems to be different for the different datasets. Camelyon17 benefits most from selecting the best model in he batch and actually has lower performance with 5 models, while FMoW (and iWildCam to a lesser degree) has a consistent increase on most metrics with more models. The relationship between between task complexity and improvement from ensembling should be studied further to understand this relationship.

\subsection{Fine-tuning}
As in the results section, we break down our analysis of fine-tuning into various dimensions including pre-training algorithm, model type, and model size.

\paragraph{Pre-Training Algorithm}

\begin{figure}[ht]
    \centering
    \includegraphics[width=\textwidth]{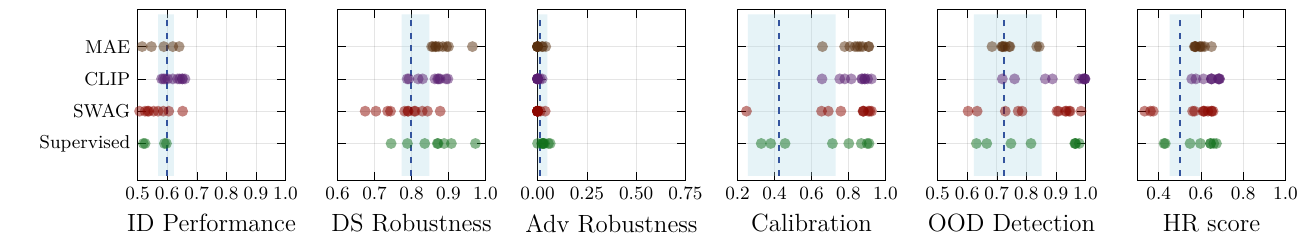}
    \caption{Effect of pre-training algorithm when fine-tuning on FMoW.}
    \label{fig:pretraining_fmow}
\end{figure}

\begin{figure}[ht]
    \centering
    \includegraphics[width=\textwidth]{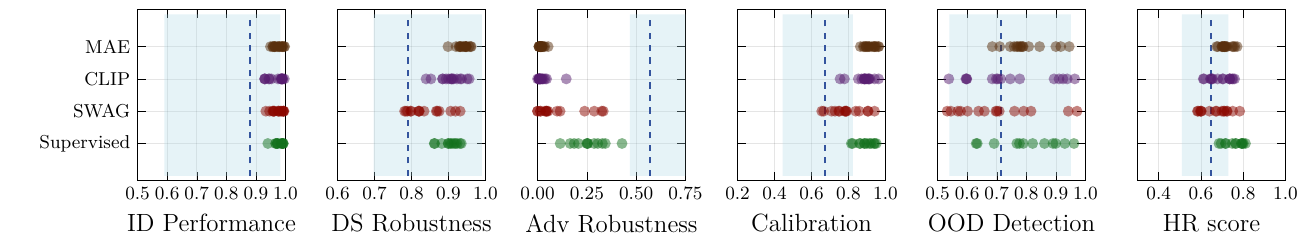}
    \caption{Effect of pre-training algorithm when fine-tuning on Camelyon17.}
    \label{fig:pretraining_camelyon17}
\end{figure}

Similar to the results form iWildCam each of the pre-training algorithms seemed to lead to a consistent improvement across all metrics except adversarial robustness for both FMoW and Camelyon17 (see \cref{fig:pretraining_fmow} and \cref{fig:pretraining_camelyon17} respectively). Interestingly, the adversarial robustness of the fine-tuned Camelyon17 models dropped significantly compared to the baseline models. We hypothesize that this is due to the large model sizes (and therefore large number of degrees of freedom) of the pre-trained models, though more investigation would be required to understand this phenomenon. 

\begin{figure}[ht]
    \centering
    \includegraphics[width=\textwidth]{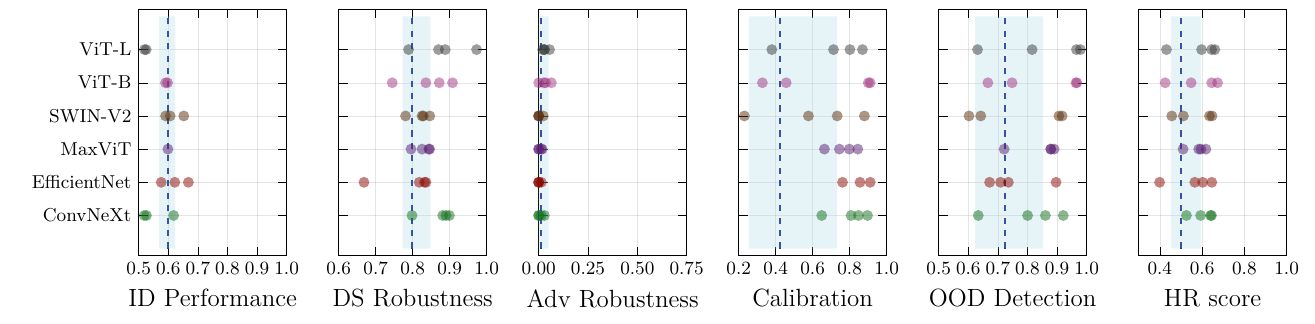}
    \caption{Effect of model architecture when fine-tuning on FMoW.}
    \label{fig:model_type_fmow}
\end{figure}

\begin{figure}[ht]
    \centering
    \includegraphics[width=\textwidth]{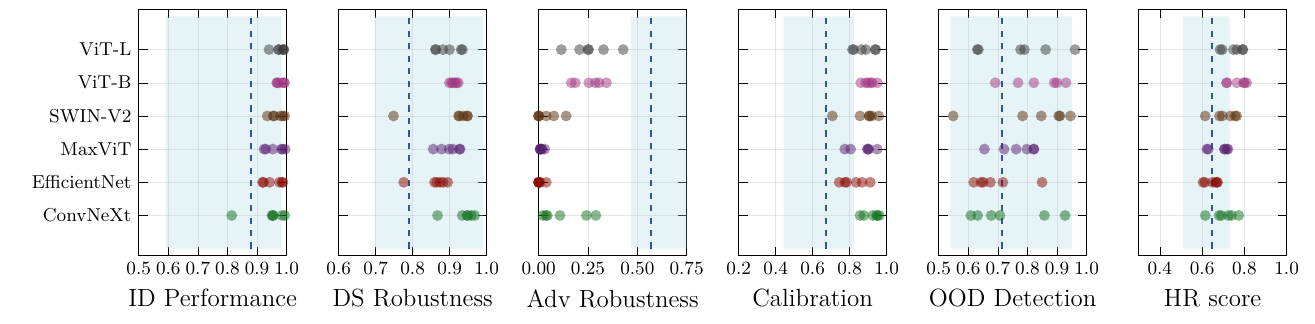}
    \caption{Effect of model architecture when fine-tuning on Camelyon17.}
    \label{fig:model_type_camelyon17}
\end{figure}

\paragraph{Model Architecture}
When comparing the results across model architectures, we find that the ViT architecture has the highest HR scores for FMoW and Camelyon17, but only by a slim margin (see \cref{fig:model_type_fmow} and \cref{fig:model_type_camelyon17} respectively)

\begin{figure}[ht]
    \centering
    \includegraphics[width=\textwidth]{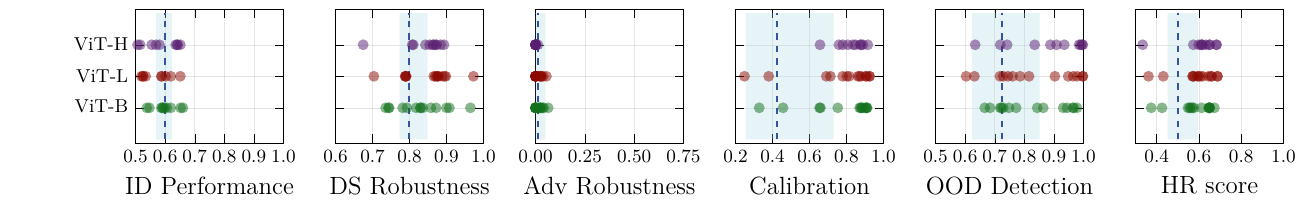}
    \caption{Effect of transformer size when fine-tuning on FMoW}
    \label{fig:model_size_fmow}
\end{figure}

\begin{figure}[ht]
    \centering
    \includegraphics[width=\textwidth]{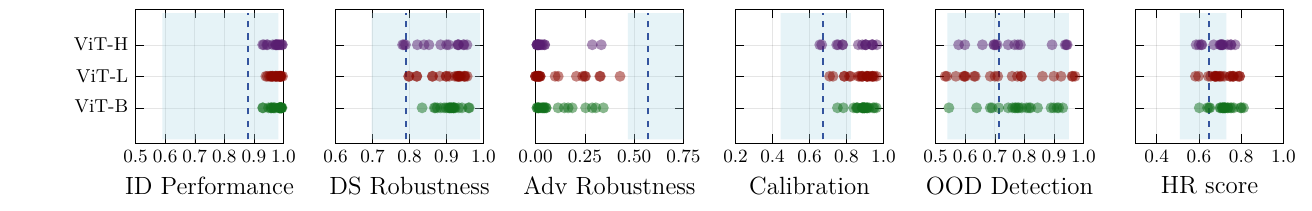}
    \caption{Effect of transformer size when fine-tuning on Camelyon17}
    \label{fig:model_size_camelyon17}
\end{figure}

\paragraph{Model Size}
Similarly to iWildCam, the trend in model size is not particularly stong with a slight improvement for larger models with FMoW (see \cref{fig:model_size_fmow}) but the opposite relationship for Camelyon17 (see \cref{fig:model_size_camelyon17}).

\subsection{Sub-Group Aware Training}
\begin{figure}[ht]
    \centering
    \includegraphics[width=\textwidth]{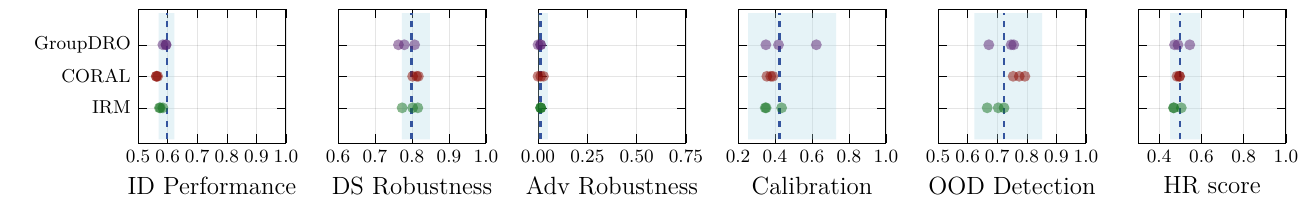}
    \caption{Effect of including sub-group information for FMoW.}
    \label{fig:invariant_loss_fmow}
\end{figure}

\begin{figure}[ht]
    \centering
    \includegraphics[width=\textwidth]{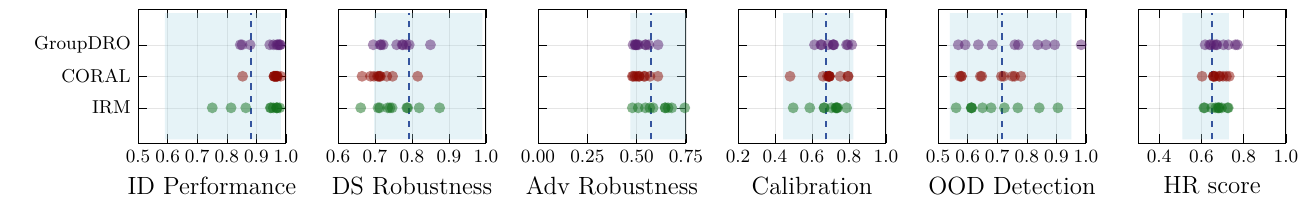}
    \caption{Effect of including sub-group information for Camelyon17.}
    \label{fig:invariant_loss_camelyon17}
\end{figure}

The algorithms for sub-group aware training do not seem to have a strong effect for any of the datasets they were trained on, leading to values within the range of the baselines. This trend is consistent for FMoW and Camelyon17 (see \cref{fig:invariant_loss_fmow} and \cref{fig:invariant_loss_camelyon17} respectively). 

\subsection{Unlabeled Training Data}
\begin{figure}[ht]
    \centering
    \includegraphics[width=\textwidth]{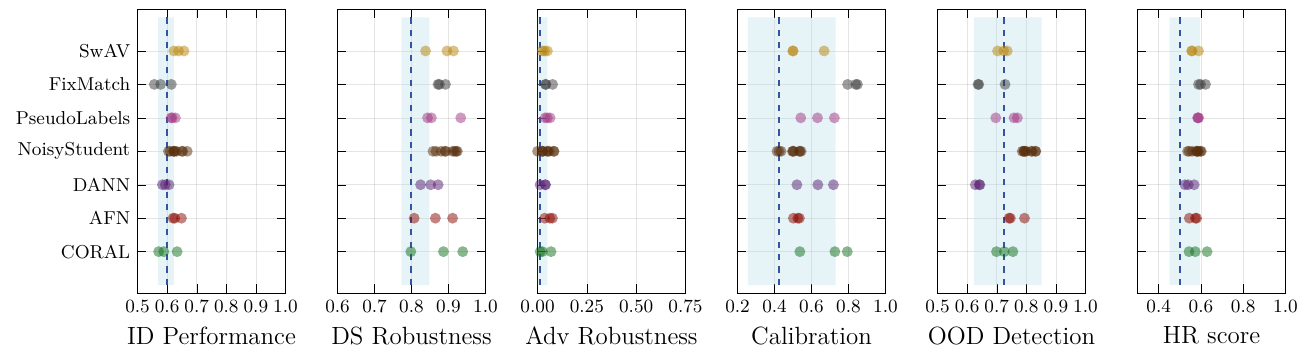}
    \caption{Effect of using unlabeled training data for FMoW.}
    \label{fig:unlabeled_data_fmow}
\end{figure}

\begin{figure}[ht]
    \centering
    \includegraphics[width=\textwidth]{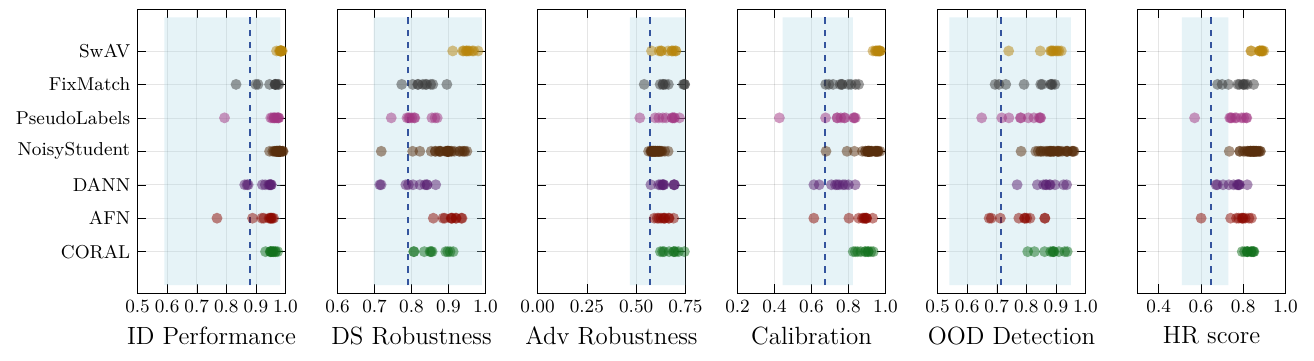}
    \caption{Effect of using unlabeled training data for Camelyon17.}
    \label{fig:unlabeled_data_camelyon17}
\end{figure}

The effect of unlabeled training data is less consistent across datasets compared to the other approaches investigated. While these approaches had a modest positive effect on the iWildCam and FMoW datasets (see \cref{fig:unlabeled_data_fmow}), they showed a larger improvement for Camelyon17 (see \cref{fig:unlabeled_data_camelyon17}). For FMoW, the unlabeled data seemed to mostly improve the DS robustness of the models (true for all algorithms), whereas for Camelyon17 it was primarily the calibration the was improved consistently. SwAV was especially effective for Camelyon17 and might indicate that if a good contrastive representation is learned from unlabeled data, it can drastically improve performance on the downstream task. 

\subsection{Distribution of Scores}

\begin{figure}[ht]
    \centering
    \includegraphics[width=\textwidth]{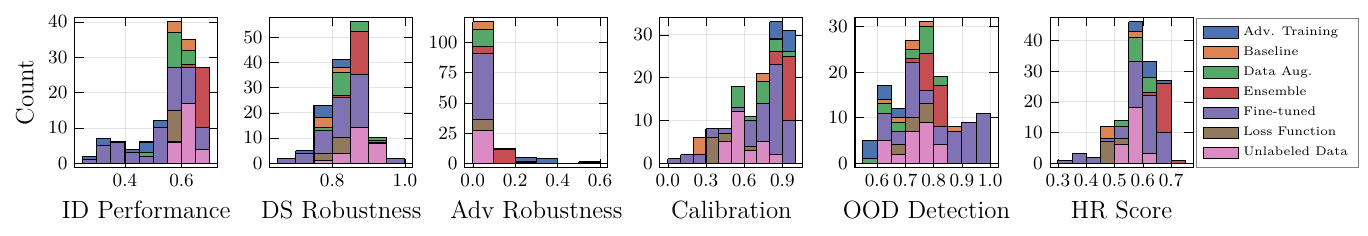}
    \caption{Distribution of scores for FMoW for various groups of models.}
    \label{fig:distribution_of_scores_fmow}
\end{figure}

\begin{figure}[ht]
    \centering
    \includegraphics[width=\textwidth]{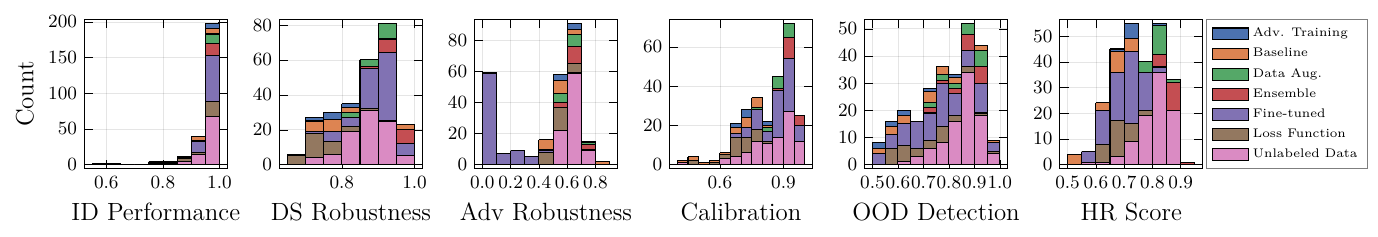}
    \caption{Distribution of scores for Camelyon17 for various groups of models.}
    \label{fig:distribution_of_scores_camelyon17}
\end{figure}

The distribution of scores stratified by groups is shown for FMoW in \cref{fig:distribution_of_scores_fmow} and for Camelyon17 in \cref{fig:distribution_of_scores_camelyon17}. 

\subsection{Metric Correlations}
\label{app:metric_correlations}

\begin{figure}[ht]
    \centering
    \includegraphics[width=\textwidth]{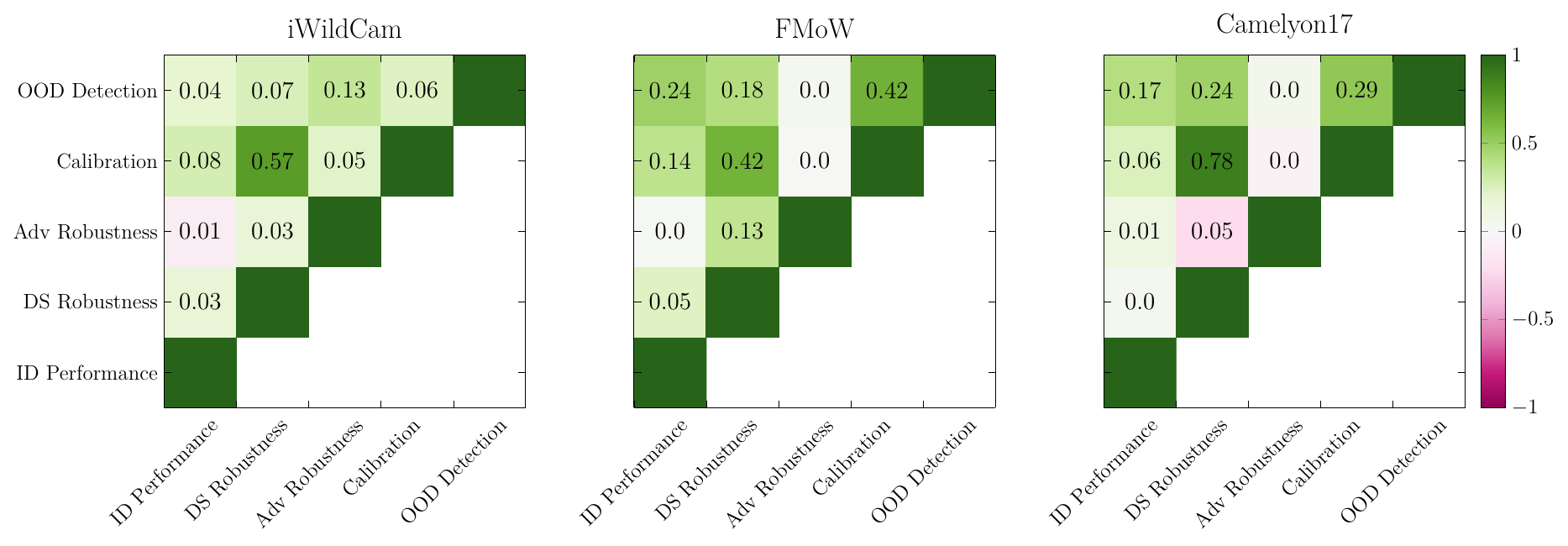}
    \caption{Correlations between reliability metrics for three datasets. Color indicates Pearson correlation and value indicates $R^2$.}
    \label{fig:metric_correlations_finetuned}
\end{figure}

While the group-mean-adjusted correlations showed that there were few correlations between metrics, there was one set of correlations that seemed consistent across datasets and that was the relationship between DS robustness and calibration. Investigating further we found that this correlation was strongest in the set of fine-tuned models. We show the metric correlations of just the fine-tuned models in \cref{fig:metric_correlations_finetuned}. We see that in this group of models the correlation is even stronger and is again consistent across datasets. We also observe that there are more positive correlation across other pairs of metrics in this group of models. This finding may suggest that some metrics become correlated when fine-tuning from large pre-trained models. It may be consistent with prior work that has uncovered a relationship between model confidence and performance on DS data~\citep{guillory2021predicting}. Additional research will need to be done to understand this relationship better. 

\end{document}